# New Movement and Transformation Principle of Fuzzy Reasoning and Its Application to Fuzzy Neural Network

CHUNG-JIN KWAK[1], SON-IL KWAK[1]*, DAE-SONG KANG[1], SONG-IL CHOE[1],
JIN-UNG KIM[1], HYOK-GI CHEA[2]

[1] College of Information Science, Kim Il Sung University, Pyongyang, D P R of Korea, **ryongnam18@yahoo.com**
[2] Pyongyang University of Computer Technology, Pyongyang, D P R of Korea, **cioc3@ryongnamsan.edu.kp**
* Corresponding Author

**Abstract**

In this paper, we propose a new fuzzy reasoning principle, so called Movement and Transformation Principle (MTP). This Principle is to obtain a new fuzzy reasoning result by Movement and Transformation the consequent fuzzy set in response to the Movement, Transformation, and Movement-Transformation operations between the antecedent fuzzy set and fuzzificated observation information. And then we presented fuzzy modus ponens and fuzzy modus tollens based on MTP. We compare proposed method with Mamdani's fuzzy system, Sugeno's fuzzy system, Wang's distance type fuzzy reasoning method and Hellendoorn's functional type method. And then we applied to the learning experiments of the fuzzy neural network based on MTP and compared it with the Sugeno's method. Through prediction experiments of fuzzy neural network on the precipitation data and security situation data, learning accuracy and time performance are clearly improved. Consequently we show that our method based on MTP is computationally simple and does not involve nonlinear operations, so it is easy to handle mathematically.

## 1. Introduction

Fuzzy set theory and fuzzy reasoning are one of the Artificial Intelligence. Since Zadeh [25] has made the conception of fuzzy sets and membership (1965), it has been widely used in various application fields, such as fuzzy control, fuzzy data mining, fuzzy expert system, and so on. The research of fuzzy reasoning aimed at more closely simulate the ability of human thinking becomes a fundamental and essential part in the fuzzy theory. So far, a number of authors have proposed various principles and methods of the fuzzy reasoning based on different ideas. The most basic fuzzy reasoning models are fuzzy modus ponens (FMP) and fuzzy modus tollens (FMT).

FMP: Given the input "$x$ is $A'$" and fuzzy rule "*if $x$ is $A$ then $y$ is $B$*", try to deduce a reasonable output "$y$ is $B'$".

FMT: Given the input "$y$ is $B'$" and fuzzy rule "*if $x$ is $A$ then $y$ is $B$*", try to deduce a reasonable output "$x$ is $A'$".

Where $A$ and $A'$ are two fuzzy sets defined on a universe X, $B$ and $B'$ are two fuzzy sets defined on another universe $Y$. The fuzzy reasoning method for solving FMT and FMP problems was first proposed by Zadeh [26], called the compositional rule of inference (CRI). In this method, a given fuzzy rule "*if $x$ is $A$ then $y$ is $B$*" is expressed as a fuzzy relation through some fuzzy implication, and the reasoning result $B'$ is calculated from the input $A'$ by using the composite operation of fuzzy relation. Although the CRI method has been successfully applied in many areas, it lacks clear logical basis and has some imperfections [21]. For example, the CRI method is not reductive [8]. In other words, if $A'$ is equal to $A$, it does not always infer that $B'$ is equal to $B$ as we expect. To improve the CRI method, Wang [20] established the Triple Implication Principle(TIP) for fuzzy reasoning and proposed the full implication triple I method based on $R_0$-implication operator. The TIP method can be considered as a complement of the CRI method, it also provides the criteria to choose the appropriate implication operator. Since the introduction of the TIP method, many studies have discussed the variants of triple I method and its applications, including reverse triple Implication method [27], triple I based on first order logical system [13], triple Implication method for interval-valued fuzzy reasoning [7], parametric triple

Implication method [7] and $\alpha$-triple Implication method [14], and so on. However, the TIP method and its variants cannot be applied in fuzzy control [3]. Mamdani [11] proposed a fuzzy reasoning method that uses the minimum operator and the sup-min composition instead of the implication operator in the classical boolean logic. Although the use of minimum operation is contrary to intuition, the Mamdani-type fuzzy reasoning has been very successfully applied in applications of fuzzy control. Takagi and Sugeno [15] proposed Takagi–Sugeno (T-S) fuzzy model, in which the antecedent consists with the fuzzy sets and the consequent is a linear functions of the fuzzy variables. This method has been widely used in fuzzy control and fuzzy identification so far. Some scholars investigated similarity-based fuzzy reasoning methods. Unlike the CRI method or triple I method, it does not require the construction of a fuzzy relation between input and output fuzzy data, and it is conceptually clearer than CRI [19]. Turksen et al. [17, 18] proposed approximate analogical reasoning schema (AARS) based on similarity measure which exhibits the advantage of fuzzy sets theory and analogical reasoning. Chen [1, 2] presented two different methods for medical diagnosis problems based on the cosine angle between the two vectors. In the study of Yeung [23, 24], the similarity measure is based on the degree of subsethood between the input information and the antecedent. They also compared and analyzed six similarity-based fuzzy reasoning methods. Wang and Meng [19] defined the fuzzy similarity measure as a generalization of the similarity measure and proposed a novel fuzzy reasoning method, called fuzzy similarity reasoning. The most commonly used fuzzy reasoning methods in applications are still CRI method and TS fuzzy model, although they are very simple and have significant disadvantages [3]. However, these methods do not explicitly use the relation between the antecedent of rule and the observation (input) in the reasoning process. For example, in the FMP model, it is natural to believe $B'$ should be close to $B$ if $A'$ is close to $A$. This means that the relation between $A$ and $A'$ should be also taken into account in the fuzzy reasoning process to get the reasoning result $B'$. Although the similarity-based fuzzy reasoning methods deduce $B'$ by modifying the consequent $B$ with a modification function based on the similarity between $A$ and $A'$, the final reasoning results strongly depend on the similarity measure and the modification function. [1]    [2] [3]

In 1992, Hans Hellendoorn proposed the generalized modus ponens (GMP) considered as a functional approach, in "Fuzzy Sets and Systems" 46(1): February (1992) 29-48. In [4], author has mentioned that the GMP is a fuzzy reasoning rule, which is as follows. A lot of the criteria for this fuzzy reasoning rule were presented, there are 3 basic assumptions for dealing with the GMP, i.e., (1) the fuzzy rule is represented by a fuzzy relation, (2) Antecedent $A$ can be strengthened or weakened to obtain new conclusion $B'$, and (3) the conclusion $B'$ is obtained by the max-min compositional rule of inference. A number of theorems have shown that these 3 assumptions are not suitable with the criteria. Furthermore, construction of an implication rule to satisfy (2) and (3) is difficult. Therefore, (3) it must be modified into some functional relation.

In this paper, we intend to develop a new fuzzy reasoning method, called Compensating Fuzzy Reasoning (MTP), which reflects the relationship between the input information and the antecedent of fuzzy rules and is consistent with human thinking. Our method considers the input information as a Movement and Transformation version of the antecedent fuzzy set. The reasoning result is obtained by applying the Movement and Transformation operations to the consequent of fuzzy rule.

This paper is organized as follows. In Section 2, we introduce the basic idea of compensating fuzzy reasoning method and formulate the FMP and FMT based on it. We also describe the logical properties of MTP method. And then we analyze about the mapping, linguistic modifier, and the role of Movement, Transformation, and Movement-Transformation operation. In Section 3, we describe the MTP method applied to two type of fuzzy system, i.e., Mamdani's and sugeno's one, and compare them with the previous methods. In section 4, we shows checking of Hellendoorn's and our method. And section 5 shows the experiment results of the proposed method through the fuzzy neural network on the precipitation data and security situation data. Finally, conclusions are drawn in Section 6.

## 2. Movement and Transformation Principle of Fuzzy Reasoning

In this Section, we describe the reasoning model of FMP and FMT, which are the most basic form of fuzzy reasoning, from a new perspective, and consider the logical properties of them.

### 2.1. Basic Idea of New Fuzzy Reasoning Principle

3. Given the fuzzy rules and the observation (input) data obtained from crisp input information, there are two kinds of relationships between the observation and the antecedent (consequent) of the fuzzy rule, i.e., Movement and Transformation. These relationships can be reflected in the antecedent (consequent) of the fuzzy rule to obtain a new consequent (antecedent) fuzzy set. We call this idea *Movement and Transformation Principle of Fuzzy Reasoning*. This is consisted of Movement method, Transformation method, and, Movement and Transformation method in fuzzy reasoning. Unlike the consequent fuzzy set obtained by using Zadeh compositional rule of inference (CRI) is a non-regular convex set, our reasoning method derives a regular convex set by the compensating operation. In general, the simplest fuzzy rule is expressed in the form of formula (1) for FMP (resp. FMT). Given the observation fuzzy set $A'$ (resp. $B'$), the process to get the fuzzy reasoning result $B'$ (resp. $A'$) can be written as formula (2) for FMP (resp. FMT).

$$\begin{aligned} FMP-Rule: &\quad if\ x\ is\ A\ then\ y\ is\ B \\ FMT-Rule: &\quad if\ y\ is\ \overline{B}\ then\ x\ is\ \overline{A} \end{aligned} \quad (1)$$

$$\begin{aligned} FMP\text{–}CFR: &\quad (A \leftrightarrow A') \Rightarrow (B \leftrightarrow B') \\ FMP\text{–}CFR: &\quad (B \leftrightarrow B') \Rightarrow (A \leftrightarrow A') \end{aligned} \quad (2)$$

where symbol "$\leftrightarrow$" indicates Movement and Transformation operation of the fuzzy set. Formula (2) implies that a new fuzzy reasoning result $B'$ (resp. $A'$) is obtained by the operations of Movement and Transformation the consequent fuzzy set $B$ (resp. $A$) in response to the Movement and Transformation relationships between the antecedent fuzzy set $A$ (resp. $B$) and the observation fuzzy set $A'$ (resp. $B'$) for FMP(resp. FMT). Note that $A'$ (resp. $B'$) is observation convex and normal fuzzy set for FMP (resp. FMT). This is obtained by Movement and Transformation based on input information $x_0$ (resp. $y_0$) for FMP(resp. FMT). The relationship between the antecedent fuzzy set and the observation includes the following three cases: the Transformation without Movement, the Movement without Transformation, and the combination of the Movement and Transformation. The type of operation is determined according to above relationship. Compositional rule of inference proposed by Zadeh has a great significance as a reasoning reflecting the human thinking. This type of reasoning is based on the principle that the fuzzy reasoning result is calculated by the composition operation according to the fuzzy implication as shown in formula (3).

$$\begin{aligned} FMP-CRI: &\quad A' \circ (A \to B) \Rightarrow B' \\ FMT-CRI: &\quad (A \to B) \circ B' \Rightarrow A' \end{aligned} \quad (3)$$

$$R = (A \to B) \quad (4)$$

As shown formula (3) and (4) the fuzzy implication $(A \rightarrow B)$ corresponds to a fuzzy relation $R$ that may be defined in several ways. The fuzzy reasoning result depends on how the fuzzy relation is defined. When the center of the observation fuzzy set expressing input information is fixed, a well-defined fuzzy relation leads to a reasoning result that is consistent with human thinking. However, if the center of observation fuzzy set changes, it is difficult to obtain meaningful reasoning result. For example, let us take look at the process of deriving the degree of ripeness from the color of the tomato. (See Fig. 1)

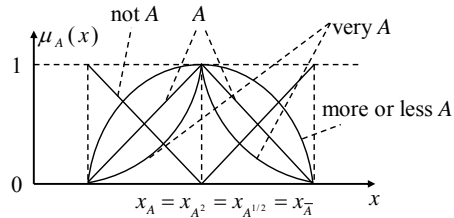

Fig. 1. Observation of fuzzy sets with same centers

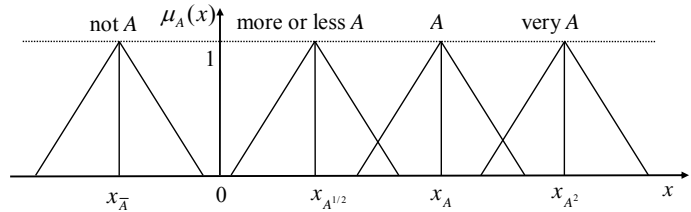

Fig. 2. Observation of fuzzy sets with different centers

Fuzzy Rule:   If a tomato is $A$ (red) then the tomato is $B$ (ripe)
Observation:   This tomato is $A'$
_______________________________________
Conclusion:   This tomato is $B'$

It is desirable that when the observation $A'$ takes "red $A$", "very red $A = A^2$", "more or less red $A = A^{1/2}$", and "not red $A = 1 - A$", the fuzzy reasoning result $B'$ is obtained as "ripe $B$", "very ripe $B = B^2$", "more or less ripe $B = B^{1/2}$", and "not ripe $B = 1 - B$", respectively. For observation fuzzy sets whose center does not change, as shown in Fig. 1, the fuzzy reasoning based on the compositional rule of inference gives us satisfactory reasoning results.

In the real world, it is more natural to express fuzzy sets from the viewpoint of the change of light wave length of the tomato color. That is, it is preferable that different observations are expressed by Movement and Transformation the centers of the fuzzy sets on the axis in the horizontal direction. (See Fig. 2) Nonetheless, fuzzy reasoning based on the compositional rules does not yield meaningful results for observation fuzzy sets with different centers. Therefore, we propose a fuzzy reasoning method based on new perspective that reasoning result $B'$ can be derived as a result of Movement or Transformation consequent fuzzy set B, because the observation $A'$ is a Movement or Transformation version of the input fuzzy set $A$. The diagrammatic explanation of CRI and MTP is shown in Fig. 3.

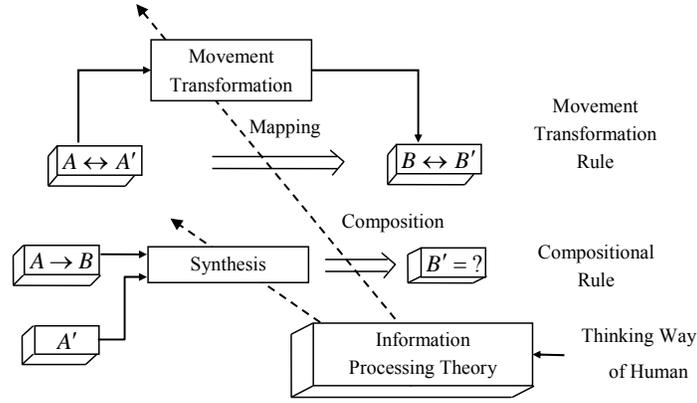

Fig. 3. Zadeh's CRI and proposed MTP

**2.2. Fuzzy Modus Ponens and Fuzzy Modus Tollens based on MTP**

**2.2.1. Fuzzy Modus Ponens based on MTP**

The fuzzy reasoning scheme of the FMP based on MTP is as follows:

| Fuzzy Rule: | if $x$ is $A$ then $y$ is $B$ |
| --- | --- |
| Observation: | $x$ is $A'$ |
| Conclusion: | $y$ is $B'$ |

**Step 1**: For the given observation fuzzy set $A'$ obtained by crisp input information $x_0 \in X$, determine the Movement and Transformation operation $F_X$ between the antecedent fuzzy set A and the observation $A'$.

$$\mu_{A'}(x) = F_X(\mu_A(x)) \qquad (5)$$

where $A, A' \in F(X)$ are fuzzy set defined on the universe of discourse $X$ and $x \in X$ is antecedent variable, $F(X)$ is the set of all fuzzy sets on $X$, and $F_X$ is a Movement, Transformation, and Movement-Transformation operation applied to the antecedent fuzzy set $A$ to obtain the given Premise $A'$ from the crisp input $x \in X$.

**Step 2**: Calculate the fuzzy reasoning result $B'$ by applying the Movement-Transformation operation $F_Y$ to the consequent fuzzy set $B$.

$$\mu_{B'}(y) = \begin{cases} F_Y(\mu_B(y)), & A \cap A' \neq \phi \\ 0, & A \cap A' = \phi \end{cases} \qquad (6)$$

where $B, B' \in F(Y)$ are fuzzy set defined on the universe of discourse $Y$, $y \in Y$ is consequent variable, $F(Y)$ is the set of all fuzzy sets on $Y$, and $F_Y$ is a Movement, Transformation, and Movement-Transformation operation applied to the consequent fuzzy set $B$ to obtain the reasoning result $B'$. Note that both $F_X$ and $F_Y$ are the projections between the fuzzy sets. The Movement-Transformation operation $F_Y$ may be defined in correspondence with the Movement-Transformation operation $F_X$ in the antecedent part, which performs the compensating action in the consequent part. The reasoning results depend on how $F_Y$ is defined. In practical applications, the Movement-Transformation operation $F_Y$ can be defined in accordance with the actual situation, which makes the reasoning method valuable and flexible.

The FMP based on MTP has an important characteristic, which is as follows.

$$A' \cap A \neq \phi \Rightarrow B' \neq \phi, \quad A' \cap A = \phi \Rightarrow B' = \phi \tag{7}$$

This means that if the observation does not match the antecedent of the fuzzy rule at all, it ignores the Movement-Transformation relationship in the antecedent and does not yield a valid reasoning result. The role of $F_X$ and $F_Y$ is considered in the subsection 2.5.

**2.2.2. Fuzzy Modus Tollens based on MTP**

The FMT based on MTP is also opposite to FMP and is represented as follows.

Fuzzy Rule: *if y is $\overline{B}$ then x is $\overline{A}$*
Observation: *y is B'*
Conclusion: *x is A'*

**Step 1**: For the given observation fuzzy set $B'$, determine the Movement-Transformation operation $F_Y$ between the consequent fuzzy set $B$ and the observation $B'$, where $B, B' \in F(Y)$.

$$\mu_{B'}(y) = F_Y(\mu_B(y)) \tag{8}$$

**Step 2**: Calculate the fuzzy reasoning result $A'$ by applying the Movement- Transformation operation $F_X$ to the antecedent fuzzy set $A$, where $A, A' \in F(X)$.

$$\mu_{A'}(x) = \begin{cases} F_X(\mu_A(y)), & B \cap B' \neq \phi \\ 0, & B \cap B' = \phi \end{cases} \tag{9}$$

In formula (8) and formula (9), the meaning of each symbol is the same as in the FMP. The Movement, Transformation, and Movement-Transformation operation $F_X$ performs the compensating action in the antecedent part. The role of Movement- Transformation operation $F_X$ and $F_Y$ is considered in the subsection 2.5. Similar to formula (7), the following fact holds also in the FMT.

$$B' \cap B \neq \phi \Rightarrow A' \neq \phi \quad or \quad B' = \phi \Rightarrow A' = \phi \tag{10}$$

Let us consider the reasoning process in detail about the FMT. Assume that the fuzzy sets $A, A' \in F(X)$ are normal and convex, and the following relation holds between them.

$$A' = \int_{A' \in F(X)} \mu_{(A')^\alpha}(x)/x = \int_{A \in F(X)} \mu_{A^\alpha}(x)/(x + \Delta x) \tag{11}$$

We call the quantities $\Delta x$ and $\alpha$ as the Movement amount and the Transformation index of the antecedent, respectively. Assume that $B, B' \in F(Y)$ are also normal and convex. Then, the fuzzy set $B'$ of fuzzy reasoning result is gained by a Movement-Transformation operation on the consequent fuzzy set $B$, for FMP, as shown in formula (12).

$$B' = \int_{B' \in F(Y)} \mu_{(B')^\beta}(y)/y = \int_{B \in F(Y)} \mu_{B^\beta}(y)/(y + \Delta y) \tag{12}$$

where $\Delta y$ and $\beta$ are the Movement amount and the Transformation index of the consequent, which are determined from $\Delta x$ and $\alpha$, respectively. In the subsection 2.5, Transformation index $\alpha$ and $\beta$ for the FMP and FMT is considered. The Movement amount of the consequent $\Delta y$ is calculated from $\Delta x$ as follows.

$$\Delta y = \begin{cases} 0, & \Delta x = 0 \\ f(\Delta x), & \Delta x \neq 0 \end{cases} \qquad (13)$$

Where $f: X \to Y$ is the mapping defined according to the relationship between the universes of discourses $X$ and $Y$. In general, mapping $f$ is set so that Movement amount $\Delta x$ and $\Delta y$ have a directly (or inverse) proportional or nonlinear relationship. In the subsection 2.5.1, we analyze for the mapping $f$.

**2.3. Logical Properties of New MTP**

In this subsection we consider logical property of qualitative criteria in the fuzzy reasoning method presented above. If any reasoning method is useful, it should satisfy the qualitative criteria that are consistent with human thinking. Mizumoto and Zimmermann [10] compared the fuzzy reasoning methods using qualitative criteria as shown in Table 1 for FMP and Table 2 for FMT. Our proposed method based on MTP satisfies the qualitative criteria of the fuzzy reasoning.

**Theorem 1.** The FMP based on MTP satisfies the qualitative criteria of the fuzzy reasoning given in Table 1.

**Table 1** Qualitative criteria for FMP

| Criteria | Observation $A'$ | Conclusion $B'$ |
| --- | --- | --- |
| criterion 1 | $x$ is $A$ | $y$ is $B$ |
| criterion 2 | $x$ is very $A$ | $y$ is very $B$ or $y$ is $B$ |
| criterion 3 | $x$ is more or less $A$ | $y$ is more or less $B$ or $y$ is $B$ |
| criterion 4 | $x$ is not $A$ | $y$ is not $B$ or $y$ is unknown |

**Proof.** Using formula (11) and (12), we can calculate the fuzzy reasoning result $B'$ for a given observation fuzzy set $A'$. In the case of $A' = A$, we know that $\alpha = 1$ and $\Delta x = 0$ since $A' = \int_{A \in F(X)} \mu_A(x)/x$. So it is obvious that $\beta = 1$ and $\Delta y = 0$, therefore the fuzzy reasoning result is $B' = \int_{B' \in F(Y)} \mu_{(B')^\beta}(y)/y = \int_{A \in F(X)} \mu_{B^\beta}(y)/(y + \Delta y) = \int_{B \in F(Y)} \mu_B(y)/y = B$.

Thus criterion 1 in Table 1 is satisfied. In Table 1, criterion 2 requires that either "$y$ is very $B$" or "$y$ is $B$" should be obtained as consequent from observation "$x$ is very $A$". In our proof, "$y$ is very $B$" is obtained, so though $y$ is $B$, criterion 2 is satisfied. In case of observation "$x$ is more or less $A$", if either "$y$ is more or less $B$" or "$y$ is $B$" is obtained, then criterion 3 is satisfied. In our method, when $\alpha = 1/2, \Delta x = 0$, then $\beta = 1/2, \Delta y = 0$. Thus the fuzzy reasoning conclusion $B'$ is obtained as follows.

$$B' = \int_{B' \in F(Y)} \mu_{B^\beta}(y)/(y + \Delta y) = \int_{B \in F(Y)} \mu_{B^{1/2}}(y)/y = \sqrt{B} = \text{more or less } B.$$

From this, though "$y$ is $B$", criterion 3 is satisfied. Lastly, in case of observation "$x$ is not $A$", Movement amount $\Delta x = 0$, and Transformation index $\alpha = not$, so $A' = \int \mu_{A^\alpha}(x)/x = \int \mu_{A^{not}}(x)/x = \int \mu_{\bar{A}}(x)/x$. And the Transformation index $\beta = \alpha = not$, Movement amount $\Delta y = \Delta x = 0$, therefore, the fuzzy reasoning result is obtained as follows.

$$B' = \int_{B' \in F(Y)} \mu_{B^\beta}(y)/(y + \Delta y) = \int_{B' \in F(Y)} \mu_{\bar{B}}(y)/(y + \Delta y) = \int_{B \in F(Y)} \mu_{\bar{B}}(y)/(y) = \text{not } B$$

In case that $A'$ is not $A$, if either " $y$ is not $B$ " or " $y$ is unknown " is obtained, qualitative criteria must be satisfied. According to our method, " $y$ is not $B$ " is obtained, so criterion 4 is satisfied. Consequently, Theorem 1 satisfies qualitative criteria for FMP. Thus this Theorem 1 is true. □

**Theorem 2.** The FMT based on MTP satisfies the qualitative criteria of the fuzzy reasoning given in Table 2.

**Table 2**  Qualitative criteria for FMT

| Criteria | Observation $B'$ | Conclusion $A'$ |
|---|---|---|
| criterion 1 | $y$ is not $B$ | $x$ is not $A$ |
| criterion 2 | $y$ is not very $B$ | $x$ is not $A$ or $x$ is not very $A$ |
| criterion 3 | $y$ is not more or less $B$ | $x$ is not $A$ or $x$ is not more or less $A$ |
| criterion 4 | $y$ is $B$ | $x$ is $A$ or $x$ is unknown |

As known in the seconds of the formula (1) and (2), FMT is opposite to FMP. Thus we can easily get the proof of Theorem 2 similar to Theorem 1. Its proof is abbreviated here.

Now let us consider the logic validness of the MTP method based on our Theorems. First, the MTP method is based on the deductive reasoning like in the FMP and the FMT based on CRI. (See Fig. 4)

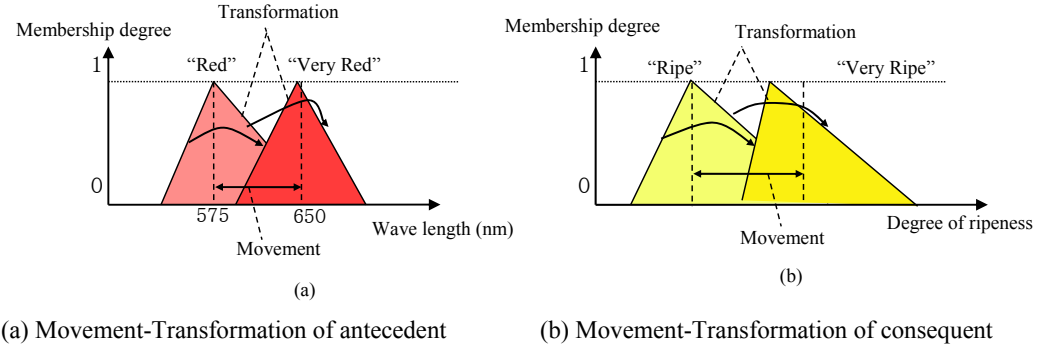

(a) Movement-Transformation of antecedent    (b) Movement-Transformation of consequent

Fig. 4. Example of fuzzy reasoning method based on MTP

In the CRI method, when a fuzzy system consisting of fuzzy rules is given and an input is entered into the system, the new knowledge (reasoning result) is derived by using the form of the logic positive and negative thought. Here, the fuzzy reasoning result is determined by how to express the relationship between the input and output of the fuzzy system, and, the way of expressing this relationship distinguishes each reasoning method. For the FMP based on the MTP, from the viewpoint that the input fuzzy set is the result of the Movement and the Transformation of the antecedent fuzzy set, the fuzzy reasoning result is also referred to as the Movement and the Transformation of the consequent fuzzy set. That is, the relationship between the input and the antecedent fuzzy set is deduced to that between the reasoning result and the consequent fuzzy set. Second, the MTP method is based on the analogy reasoning. In the MTP method, the Movement amount of the consequent has a directly (or inverse) relationship with that of the antecedent. Moreover, the fuzzy reasoning result obtained by the Transformation of the consequent has a similarity with the input fuzzy set. This means that Movement-Transformation operation in the consequent of the fuzzy rule belongs to the form of the analogy reasoning.

Therefore, it is possible to say that the MTP is a reasoning method embodying the human thinking way that combines the deductive reasoning and the analogy reasoning.

Let us consider a simple and typical example of fuzzy reasoning method based on MTP .

| Fuzzy Rule: | if a tomato is red then the tomato is ripe |
|---|---|
| Observation: | This tomato is very red |
| Conclusion: | This tomato is very ripe |

In Fig. 4, the conclusion "tomato is very ripe" was obtained by reflecting of the Movement-Transformation relationship between the antecedent fuzzy set "red" of the fuzzy rule "if a tomato is red then the tomato is ripe" and the observation fuzzy set "very red" in the consequent part. Hence, there is the similarity between the fuzzy set "very red" in the antecedent and the fuzzy set "very ripe" in the consequent.

### 2.4. Computational Example for Different Input Information

In this subsection, we analyze the fuzzy reasoning results obtained by the FMP based on the MTP when the input information is placed on the left and right hand in the center of the antecedent fuzzy set. (See Fig. 5)

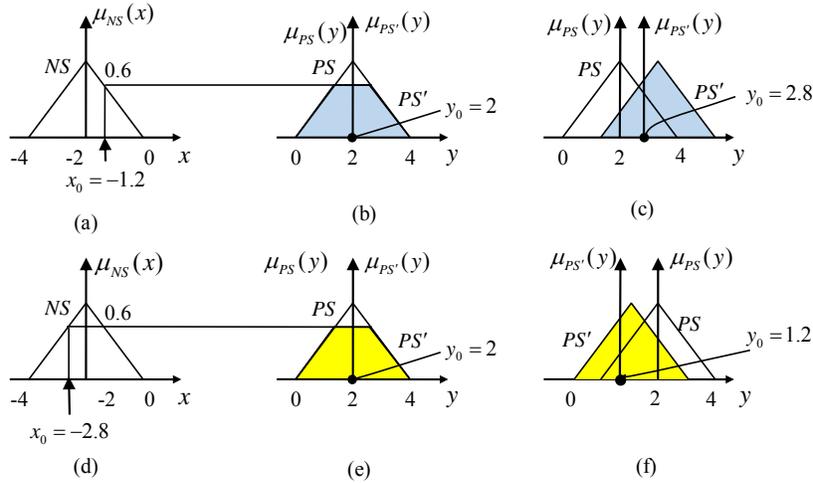

Fig. 5. Fuzzy reasoning based on MTP for different input information

Assume that the following fuzzy control rule is given: "if a temperature deviation $x$ of the reaction tank is negative small (NS), then turn the motor valve $y$ slightly clockwise (PS)". Let us consider the reasoning result of Mamdani-type fuzzy reasoning method. As shown in Fig. 5(a), if the input information $x_0$ (temperature deviation) is $-1.2$, then $\mu_{NS}(x_0) = \mu_{NS}(-1.2) = 0.6$. Therefore, the reasoning result $PS'$ is the shaded area of Fig. 5(b) and the defuzzified crisp value of the output $y_0$ (i.e., opening degree of motor valve) is 2. For the crisp input information $x_0 = -2.8$, since $\mu_{NS}(x_0) = \mu_{NS}(-2.8) = 0.6$, the fuzzy reasoning result $PS'$ is the shaded area of Fig. 5(e) and its defuzzified crisp value $y_0$ is also 2. In this case, the Mamdani-type fuzzy reasoning method gives the same reasoning result for different input. However, the MTP method differs from it. When the crisp input information $x_0$ is $-1.2$, then , the MTP method gives the crisp output reasoning result $y_0 = y_{PS} + \Delta y = y_{PS} + \Delta x = 2 + 0.8 = 2.8$, and if crisp input information $x_0 = -2.8$, then $y_0 = y_{PS} + \Delta y = 2 - \Delta x = 2 - 0.8 = 1.2$. (See Fig. 5(c) and Fig. 5(f)). That is, different inputs yield different results. In the Mamdani-type fuzzy reasoning method, the same reasoning results are always obtained for arbitrary inputs belonging to the interval [−4, 0]. Using the MTP method, in contrast, leveled by −2, the bigger the input information, the bigger the defuzzified crisp value of $PS'$, and the smaller the input information, the smaller it becomes.

In general, it is true that the several rules work to get the reasoning result for the given input information of any time in fuzzy control or fuzzy expert system. Since the reasoning result is obtained from several rules, the logical contradiction that different inputs have the same result for one rule was not questioned. However, it is more reasonable that different results are obtained for different inputs. Our MTP method is an improved general fuzzy reasoning method by which the different reasoning results are obtained from different input information.

### 2.5. Analysis of New Fuzzy Reasoning Method

In this section we analyze about mapping $f$ and, Transformation index $\alpha$ and $\beta$, and the role of Movement, Transformation, and Movement-Transformation operation presented in subsection 2.2 of this paper.

#### 2.5.1. Analysis of Movement Amount $\Delta x$ and $\Delta y$ by Mapping $f$

Let us analyze the mapping $f : X \to Y$ in formula (13). In the case $\Delta x \neq 0$ in formula (13), $\Delta y = f(\Delta x)$ may be illustratively defined as follows.

$$\Delta y = f(\Delta x) = k \cdot \Delta x \tag{14}$$

, where $\Delta y$ and $\Delta x$ are the same as mentioned above, and k is the proportional coefficient in the closed interval $[0, l]$, and $l$ is the maximum of $\Delta x$. For example, in case $k = 0.3, 1, 1.2$, $\Delta y = 0.3\Delta x$, $\Delta y = 1 \cdot \Delta x$, $\Delta y = 1.5\Delta x$. In the practical application, $k$ is determined by designer or engineer according to the characteristics of the object and spot experience. Generally if $k > 1$, Movement amount $\Delta y$ is increased, $k = 1$, $\Delta y = \Delta x$, and if $0 < k < 1$, it is decremented.

Other definition of $\Delta y$ can be illustratively described as follows.

$$\Delta y = f(\Delta x) = (\Delta x)^k \tag{15}$$

, where $k$ means increase or decrease of $\Delta x$. For example, in case $k = 0, 1, 2, 3$, $\Delta y = 1$, $\Delta y = \Delta x$, $\Delta y = (\Delta x)^2$, and $\Delta y = (\Delta x)^3$. In case $k = 0$, $\Delta y = 1$ means the Movement amount of consequent is 1 though the antecedent fuzzy set has Movement amount $\Delta x$. From this definition, it can be Shown that in case $0 < \Delta x < 1$, $\Delta y$ decreases according to the increase of $k$, and in case $\Delta x > 1$, $\Delta y$ increases according to the increase of $k$. From the above 2 definitions, $k$ can be determined with the experimental method. The definition of mapping $f$ can be made variously according to the characteristics of system besides the above 2 methods. The above definition of Movement amount may be applied to several branches such as water level control of water tank, stabilization control of inverted equilibrium, image processing, expert system, temperature prediction of furnace, medical diagnosis, and pattern recognition, and so on.

#### 2.5.2. Analysis of the Transformation index $\alpha$ and $\beta$ for FMP and FMT

Let us analyze the Transformation index $\alpha$ and $\beta$ in the FMP and FMT.

In fact, $\alpha$ of formula (11) is the Transformation index or hedge for the observation fuzzy set $A'$, while $\beta$ of formula (12) is the that for the reasoning result set $B'$. Generally, it is taken as $\beta \neq \alpha$. If the antecedent fuzzy set $A$ and consequent fuzzy set $B$ is scaled in the closed interval, then we can consider as $\beta = \alpha$ for FMP. The fuzzy reasoning result $B'$ is determined only by the Movement operation when $\alpha = 1$ and $\Delta x \neq 0$, and is determined only by the Transformation operation when $\alpha \neq 1$ and $\Delta x = 0$. On the other hand, if $\alpha \neq 1$ and $\Delta x \neq 0$, the Movement and Transformation operations can be applied together to obtain the fuzzy reasoning result. Consider the relation between $\alpha$ and $\beta$ in FMP. In formula (11), $\alpha$ is the Transformation index of the antecedent fuzzy set $A$. In other words, observation fuzzy set $A'$ is obtained by the Transformation of antecedent fuzzy set $A$ by using of $\alpha$. Let $\beta = m \cdot \alpha$ for FMP, where $m$ is called proportional coefficient. For example. If $m = 0, 1, 2$ then the Transformation index $\beta = 0$, $\beta = \alpha$ and $\beta = 2\alpha$, therefore the new consequent fuzzy set $B'$ in FMP, that is, fuzzy reasoning result is obtained variously as follows, respectively.

$$B' = \int_{B' \in F(Y)} \mu_{(B')^0}(y)/y = \int_{B \in F(Y)} 1/(y + \Delta y)$$

$$B' = \int_{B \in F(Y)} \mu_{B^\beta}(y)/y = \int_{B \in F(Y)} \mu_{B^\alpha}(y)/(y + \Delta y)$$

$$B' = \int_{B \in F(Y)} \mu_{B^\beta}(y)/y = \int_{B \in F(Y)} \mu_{B^{2\alpha}}(y)/(y + \Delta y)$$

Then, the relation between the Transformation index $\beta$ and $\alpha$ in FMT is similar to in FMP, so it is omitted here. In general, the Transformation index $\alpha \neq \beta$, it can be illustratively defined as $\alpha = m \cdot \beta$ for FMT, where $m$ is also called proportional coefficient. Therefore the designer should determine the reasonable $\alpha$ according to the characteristics of the object. Consequently in FMP and FMT, in case that fuzzy set $A, B,$ and $A'$ (resp. $B'$) of the fuzzy rule and antecedent(the given premise) are scaled in closed unit interval, it can be simply treated as $\beta = \alpha$ (resp. $\alpha = \beta$) for FMP(resp. FMT).

### 2.5.3. Analysis of the Role of Movement-Transformation Operation

Let us analyze the role of Movement-Transformation operation $F_X$ and $F_Y$. Now, consider the formula (5), (6), (8) and (9). Formula (5) and (6) is applied in FMP, and formula (8) and (9) in FMT.

First, consider the role of $F_X$ for FMP in formula (5). In formula (5), observation fuzzy set $A'$ is obtained by Transformation, Movement, or Movement-Transformation of antecedent fuzzy set $A$. That is, by the fuzzy set theory in paper [23], observation fuzzy set $A'$ corresponds to the fuzzification of crisp input information. In formula (2), $A \leftrightarrow A'$ can be written as $A(x) \leftrightarrow A'(x)$ in detail. Consider the role of $F_X$ in case that $A$ moves to $A'$.

**Proposition 1.** Suppose that input information $x_0 \in X$ is away from the center $x_A$ of triangular-shaped fuzzy set $A$

as $\Delta x$. When $x_0$ is crisp input information, $A'$ can be obtained by Movement $A$ as $\Delta x = |x_0 - x_A|$.

**Proof.** For this proof, role of $F_X$ and $F_Y$ in case of Movement for FMP is illustratively shown in Fig. 6.

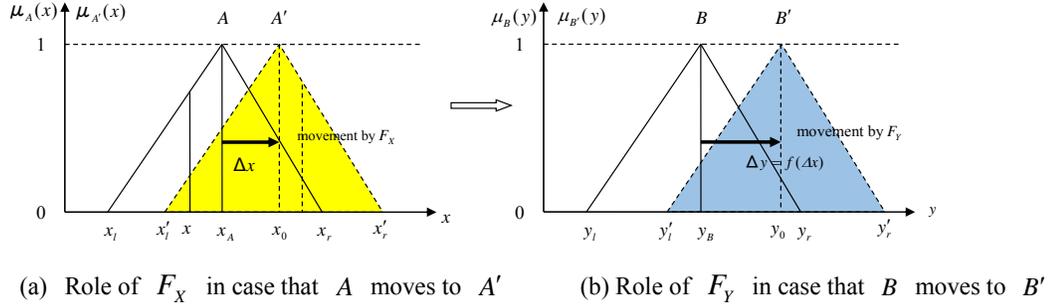

(a) Role of $F_X$ in case that $A$ moves to $A'$      (b) Role of $F_Y$ in case that $B$ moves to $B'$

Fig. 6. Role of $F_X$ and $F_Y$ in case of Movement for FMP

In Fig. 6, $x_A$ is the coordinate value of the center of triangular-shaped fuzzy set $A$, and $x_0$ is the coordinate value of new input information. On the basis of the idea in subsection 2.1, it can be Shown that $A$ moves to $A'$ linearly. Using the distance between $x_0$ and $x_A$, and left width $w_{xl}$, right width $w_{xr}$ of $A$, from formula (5) $A'$ can be written as follows.

$$\mu_{A'}(x) = F_X(\mu_A(x)) = \int_{A \in F(X)} \mu_A(x)/(x+\Delta x) = \int_{A' \in F(X)} \mu_{A'}(x)/x_0 = \int_{x \in [x'_l, x_0]} (\frac{x - x'_l}{x_0 - x'_l})/x \ \cup \int_{x \in [x_0, x'_r]} (\frac{x'_r - x}{x'_r - x_0})/x$$

$$= \int_{x \in [x_l, x_0]} (1 + \frac{x - x_0}{w_{xl}})/(x+\Delta x) \ \cup \int_{x \in [x_0, x_r]} (1 + \frac{x_0 - x}{w_{xr}})/(x+\Delta x) \quad (16)$$

Therefore we can know that this proposition 1 is true. □

**Proposition 2.** For triangular-shaped fuzzy set $A'$ obtained by the formula (16), fuzzy reasoning result $B'$ is concluded by formula (17).

$$\mu_{B'}(y)) = \int_{y \in [y_l, y_B]} (1 + \frac{y - y_B - \Delta x}{w_{yl}})/(y + \Delta x) \cup \int_{y \in [y_B, y_r]} (1 + \frac{y_B - y + \Delta x}{w_{yr}})/(y + \Delta x) \quad (17)$$

**Proof.** Using the distance between $y_0$ and $y_B$, and left width $w_{yl}$, right width $w_{yr}$ of triangular-shaped fuzzy set $B$, from formula (6), the fuzzy reasoning result $B'$ can be written as follows.

$$\mu_{B'}(y) = F_Y(\mu_B(y)) = \int_{B \in F(Y)} \mu_B(y)/(y+\Delta y) = \int_{B' \in F(Y)} \mu_{B'}(y)/y$$

$$= \int_{y \in [y'_l, y_0]} (\frac{y - y'_l}{y_0 - y'_l})/y \ \cup \ \int_{y \in [y_0, y'_r]} (\frac{y'_r - y}{y'_r - y_0})/y$$

$$= \int_{y \in [y_l, y_B]} (1 + \frac{y - y_B - \Delta y}{w_{yl}})/(y + \Delta y) \ \cup \ \int_{y \in [y_B, y_r]} (1 + \frac{y_B - y + \Delta y}{w_{yr}})/(y + \Delta y)$$

$$= \int_{y \in [y_l, y_B]} (1 + \frac{y - y_B - \Delta x}{w_{yl}})/(y + \Delta x) \ \cup \ \int_{y \in [y_B, y_r]} (1 + \frac{y_B - y + \Delta x}{w_{yr}})/(y + \Delta x)$$

From the above formula (16) and (17), it can be shown that $F_X$ plays a role that moves fuzzy set $A$ to $A'$ and $F_Y$ plays a role that moves fuzzy set $B$ to $B'$. So we can see that this proposition 2 is true.

Next, let us consider the role of $F_X$ in formula (5) in the case that observation $A$ is Transformation to fuzzy set $A'$ obtained by crisp input $x_0$ as shown in Fig. 7 (a).

**Proposition 3.** For the given premise $A' = very\ A$, Transformation fuzzy set $A'$ is described by formula (18).

$$\mu_{A'}(x) = \int_{x \in [x_l, x_0]} (1 + \frac{x - x_0}{w_{xl}})^2 /(x + \Delta x) \ \cup \ \int_{x \in [x_0, x_r]} (1 + \frac{x_0 - x}{w_{xr}})^2 /(x + \Delta x) \qquad (18)$$

**Proof.** For this proof, we show the role of $F_X$ in case that triangular-shaped fuzzy set $A$ is Transformation to fuzzy set $A'$ .(See Fig. 7) Fig. 7 (a) shows the case that observation fuzzy set $A'$ can be obtained by applying linguistic modifier very to triangular-shaped fuzzy set $A$ of fuzzy rule.

$$\mu_{A'}(x) = F_X(\mu_A(x)) = \int_{A' \in F(X)} \mu_{A'}(x)/x = \int_{x \in [x_l, x_0]} \mu_{A'}(x)/x \ \cup \ \int_{x \in [x_0, x_r]} \mu_{A'}(x)/x =$$

$$= \int_{x \in [x_l, x_0]} \mu_{A^2}(x)/x \ \cup \ \int_{x \in [x_0, x_r]} \mu_{A^2}(x)/x = \int_{x \in [x_l, x_0]} (\frac{x - x_l}{x_0 - x_l})^2 /x \ \cup \ \int_{x \in [x_0, x_r]} (\frac{x_r - x}{x_r - x_0})^2 /x$$

$$= \int_{x \in [x_l, x_0]} (1 + \frac{x - x_0}{w_{xl}})^2 /(x + \Delta x) \ \cup \ \int_{x \in [x_0, x_r]} (1 + \frac{x_0 - x}{w_{xr}})^2 /(x + \Delta x)$$

Therefore we can know that this proposition 3 is true. □

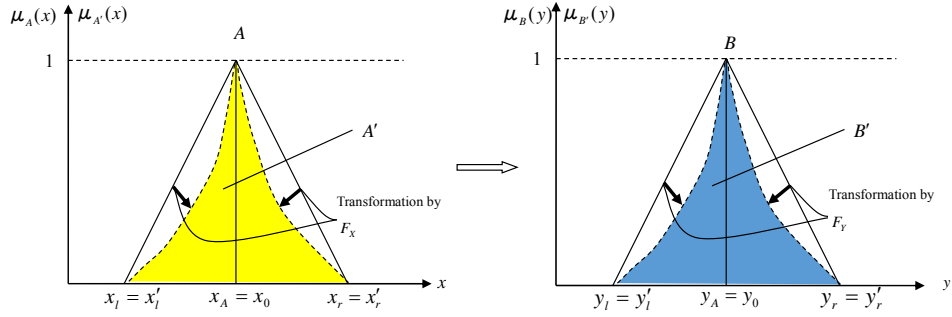

(a) Role of $F_X$ in case that $A$ is Transformation to $A'$     (b) Role of $F_Y$ in case that $B$ is Transformation to $B'$

Fig. 7. Role of $F_X$ and $F_Y$ in case of Transformation for FMP

**Proposition 4.** For the given premise $A' = very\ A$, fuzzy reasoning result, i.e., Transformation fuzzy set $B'$ is concluded by formula (19).

$$\mu_{B'}(y) = \int_{y\in[y_l,\ y_B]} (1+\frac{y-y_B-\Delta x}{w_{yl}})^2 /(y+\Delta x) \cup \int_{y\in[y_B,\ y_r]} (1+\frac{y_B-y+\Delta x}{w_{yr}})^2 /(y+\Delta x) \tag{19}$$

**Proof.** Fig.7 (b) shows the role of $F_Y$ in case that triangular-shaped fuzzy set $B$ is Transformation to Transformation fuzzy set $B'$. From Fig.7 (b) the fuzzy reasoning result $B'$ for FMP is obtained as follows.

$$\mu_{B'}(y) = F_Y(\mu_B(y)) = \int_{B'\in F(Y)} \mu_{B'}(y)/y' = \int_{y\in[y_l,y_0]} \mu_{B'}(y)/y' \cup \int_{y\in[y_0,y_r]} \mu_{B'}(y)/y' =$$

$$= \int_{y\in[y_l,y_0]} \mu_{B^2}(y)/y_0 \cup \int_{y\in[y_0,y_r]} \mu_{B^2}(y)/y_0$$

$$= \int_{y\in[y_l,y_0]} (\frac{y-y_l}{y_0-y_l})^2 /(y+\Delta y) \cup \int_{y\in[y_0,y_r]} (\frac{y_r-y}{y_r-y_0})^2 /(y+\Delta y)$$

$$= \int_{y\in[y_l,\ y_B]} (1+\frac{y-y_B-\Delta y}{w_{yl}})^2 /(y+\Delta y) \cup \int_{y\in[y_B,\ y_r]} (1+\frac{y_B-y+\Delta y}{w_{yr}})^2 /(y+\Delta y)$$

$$= \int_{y\in[y_l,\ y_B]} (1+\frac{y-y_B-\Delta x}{w_{yl}})^2 /(y+\Delta x) \cup \int_{y\in[y_B,\ y_r]} (1+\frac{y_B-y+\Delta x}{w_{yr}})^2 /(y+\Delta x)$$

Therefore, it can be shown that in case that $A$ is Transformation to $A'$, Transformation, that is, mapping $F_X$ plays a role of Transformation fuzzy set $A$ to $A'$. This corresponds to the fuzzification based on the crisp input information in general fuzzy reasoning. Thus we can see that this proposition 4 is true. □

Then, consider the role of $F_X$ in case that Transformation and Movement exist at the same time in formula (5).

**Proposition 5.** For the given premise $A' = more\ or\ less\ A$, Transformation fuzzy set $A'$ is described as formula

(20).

$$\mu_{A'}(x) = \int_{x \in [x_l, x_0]} (1 + \frac{x - x_0}{w_{xl}})^{\frac{1}{2}} /(x + \Delta x) \ \cup \ \int_{x \in [x_0, x_r]} (1 + \frac{x_0 - x}{w_{xr}})^{\frac{1}{2}} /(x + \Delta x) \tag{20}$$

**Proof.** Suppose that $A'$ is moved from triangular-shaped fuzzy set $A$ as Movement amount $\Delta x$ and Transformation by the linguistic modifier more or less. That is, $A' = more\ or\ less\ A$. (See Fig. 8) Fig. 8 (a) shows the example of the role of $F_X$ in formula (5) in case that $A$ is moved and transformed to $A'$. From Fig. 6 (a), Fig. 7 (a), and formula (5), fuzzy set $A'$ can be written as follows as shown Fig. 8 (a). From this formula, we can know that $F_X$ plays a role that deforms fuzzy set $A$ as much as linguistic modifier (hedge) "more or less" and moves as much as $\Delta x$ on the axis $x$. The formula (20) shows that the fuzzy set $A'$ is obtained by applying the Movement-Transformation operation $F_X$ to the antecedent fuzzy set $A$.

$$\mu_{A'}(x) = F_X(\mu_A(x)) = \int_{A' \in F(X)} \mu_{A'}(x)/(x + \Delta x) \ = \ \int_{x \in [x'_l, x_0]} \mu_{A'}(x)/x \ \cup \ \int_{x \in [x_0, x'_r]} \mu_{A'}(x)/x =$$

$$= \int_{x \in [x'_l, x_0]} \mu_{A^{\frac{1}{2}}}(x)/x \ \cup \ \int_{x \in [x_0, x'_r]} \mu_{A^{\frac{1}{2}}}(x)/x$$

$$= \int_{x \in [x'_l, x_0]} (\frac{x - x'_l}{x_0 - x'_l})^{\frac{1}{2}}/x \ \cup \ \int_{x \in [x_0, x'_r]} (\frac{x'_r - x}{x'_r - x_0})^{\frac{1}{2}}/x$$

$$= \int_{x \in [x_l, x_0]} (1 + \frac{x - x_0}{w_{xl}})^{\frac{1}{2}} /(x + \Delta x) \ \cup \ \int_{x \in [x_0, x_r]} (1 + \frac{x_0 - x}{w_{xr}})^{\frac{1}{2}} /(x + \Delta x)$$

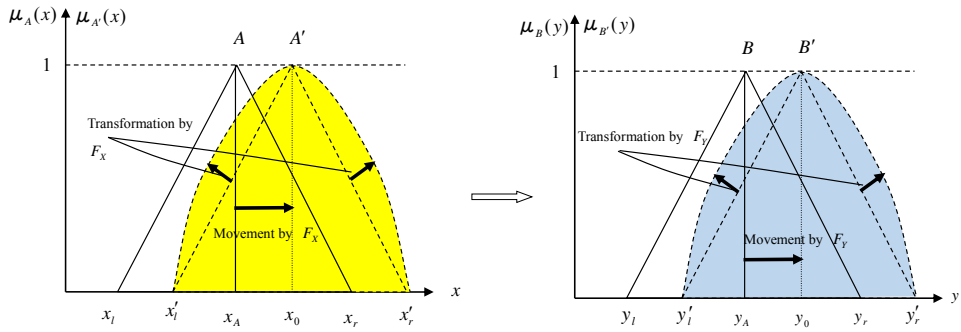

(a) Role of $F_X$ in case that $A$ is moved and transformed to $A'$ (b) Role of $F_Y$ in case that $B$ is moved and transformed to $B'$

Fig. 8. Role of $F_X$ and $F_Y$ in case of Movement and transformation for FMP

Finally, consider the role of $F_Y$ in formula (6). As result, $F_Y$ in formula (6) plays a role that moves and transforms, or Movement-Transformation consequent fuzzy set $B$ in FMP. So we can know that this proposition 5 is true. □

**Proposition 6.** For the given premise $A' = very\ A$, the fuzzy reasoning result $B'$ is concluded by formula (21).

$$\mu_{B'}(y) = \int_{y\in[y_l,\,y_B]} (1+\frac{y-y_B-\Delta x}{w_{yl}})^{\frac{1}{2}}/(y+\Delta x) \cup \int_{y\in[y_B,\,y_r]} (1+\frac{y_B-y+\Delta x}{w_{yr}})^{\frac{1}{2}}/(y+\Delta x) \qquad (21)$$

**Proof.** Let $y_B$ be the coordinate value of the center of $B$ and $y_0$ be the center of fuzzy set $B'$ of fuzzy reasoning result newly obtained. From Fig. 8 (b) the fuzzy reasoning result for FMP is obtained as follows.

$$\begin{aligned}
\mu_{B'}(y) &= F_Y(\mu_B(y)) = \int_{B'\in F(Y)} \mu_{B'}(y)/(y+\Delta y) = \int_{y\in[y_l',y_0]} \mu_{B'}(y)/y \ \cup \ \int_{y\in[y_0,y_r']} \mu_{B'}(y)/y \\
&= \int_{y\in[y_l',y_0]} \mu_{B^{\frac{1}{2}}}(y)/y \ \cup \ \int_{y\in[y_0,y_r']} \mu_{B^{\frac{1}{2}}}(y)/y = \int_{y\in[y_l',y_0]} (\frac{y-y_l'}{y_0-y_l'})^{\frac{1}{2}}/y \ \cup \ \int_{y\in[y_0,y_r']} (\frac{y_r'-y}{y_r'-y_0})^{\frac{1}{2}}/y \\
&= \int_{y\in[y_l,\,y_B]} (1+\frac{y-y_B-\Delta y}{w_{yl}})^{\frac{1}{2}}/(y+\Delta y) \ \cup \ \int_{y\in[y_B,\,y_r]} (1+\frac{y_B-y+\Delta y}{w_{yr}})^{\frac{1}{2}}/(y+\Delta y) \\
&= \int_{y\in[y_l,\,y_B]} (1+\frac{y-y_B-\Delta x}{w_{yl}})^{\frac{1}{2}}/(y+\Delta x) \ \cup \ \int_{y\in[y_B,\,y_r]} (1+\frac{y_B-y+\Delta x}{w_{yr}})^{\frac{1}{2}}/(y+\Delta x)
\end{aligned}$$

Let $y_l$ be the left endpoint of $B$, and $y_r$ the right endpoint of $B$. Then based on the idea of subsection 2.1, by the Movement of $B$, the left endpoint of $B'$ is $y_l'$, and right endpoint is $y_r'$. That is, it can be shown that new reasoning result is obtained by move-Transformation the fuzzy set $B$ to $B'$. For example, suppose that Transformation index is $\beta = \alpha = \frac{1}{2}$. The formula (21) shows that the fuzzy reasoning result $B'$ is obtained by applying the Movement-Transformation operation $F_Y$ to the consequent fuzzy set $B$ from Fig. 8. Consequently we can see that this proposition 6 is true. □

Next, let us analyze the role of $F_Y$ and $F_X$ for FMT in formula (8) and (9). This can be analyzed similar to the formula (5) and (6) for FMP. That is, $F_Y$ plays a role that obtains the observation $B'$ by Movement and Transformation, or Movement-Transformation of consequent fuzzy set $B$. This corresponds to the fuzzification of crisp input information in the general FMT. And the role of $F_X$ in formula (9) can be analyzed similar to in formula (6). That is, in formula (9), The Movement, Transformation, and Movement-Transformation operation $F_X$ plays a role that infer a new conclusion fuzzy set $A'$ from $A$ by Movement amount $\Delta y$ and Transformation of $B'$ from $B$, where $B'$ is obtained on the basis of $F_Y$ in formula (8). In this paper, since the role of $F_Y$ and $F_X$ for FMT is similar for FMP, concrete expression is omitted.

So far, we analyzed the role of $F_X$ (resp. $F_Y$) and $F_Y$ (resp. $F_X$) in FMP(resp. FMT) according to the of fuzzy reasoning based on compensating operation in case that input information fuzzy set $A'$ (resp. $B'$) is obtained by the Movement, Transformation, and Movement-Transformation of $A$ (resp. $B$).

## 4. Adaption to Several Fuzzy Systems of MTP

### 3.1. Mamdani Fuzzy System Based on MTP

In this subsection, we apply our MTP for the Mamdani fuzzy system [9], in which the consequents consist of fuzzy sets. Assume that the fuzzy system is given as formula (22).

$$
\begin{aligned}
&R_j: \text{ if } x_1 \text{ is } P_{j1}, \cdots, x_i \text{ is } P_{ji}, \cdots, x_s \text{ is } P_{js} \text{ then } z_j = Q_j \\
&\underline{\text{Input: } x_1 \text{ is } P_1, \cdots, x_i \text{ is } P_i, \cdots, x_s \text{ is } P_s} \\
&\quad \text{Conclusion: } z \text{ is } z^0 = ?
\end{aligned}
\tag{22}
$$

where $x_i \in X_i, i=1,2,\cdots,s$ is the input variable, $z \in Z$ is the output variable, $P_{ji} \in F(X_i), i=1,2,\cdots s,\ j=1,2,\cdots,n$ is the antecedent fuzzy set of $j^{th}$ rule, $Q_j \in F(Z),\ j=1,2,\cdots,n$ is the consequent fuzzy set of $j^{th}$ rule, $P_i \in F(X_i),\ i=1,2,\cdots,s$ is the $i^{th}$ input fuzzy set, $Q' \in F(Z)$ is the conclusion fuzzy set, s is the number of inputs, and n is the number of rules. Without loss of generality, we assume that all of the fuzzy sets are triangular-shaped function. (See Fig. 9)

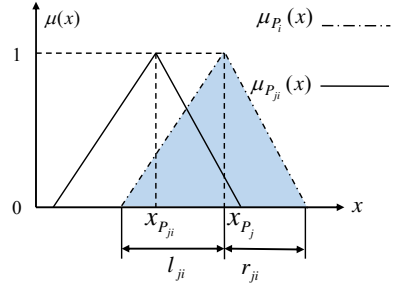

Fig. 9. Movement operation of triangular-shaped fuzzy sets

Our method consists of the following three steps:

**Step 1:** Determine the Movement amounts of antecedent for all fuzzy rules. First, calculate the Movement amount between the $i^{th}$ input triangular-shaped fuzzy set $P_i$ and antecedent triangular-shaped fuzzy set $P_{ji}$ of rule $R_j$ as follows.

$$
\Delta x_{ji} = \begin{cases} (x_{P_{ji}} - x_{P_i})/l_{ji}, & x_{P_i} \leq x_{P_{ji}} \\ (x_{P_{ji}} - x_{P_i})/r_{ji}, & x_{P_{ji}} < x_{P_i} \end{cases}, \quad i=1,2\cdots,s,\ j=1,2,\cdots,n
\tag{23}
$$

where $l_{ji}$ and $r_{ji}$ are the left width and right width of antecedent fuzzy set $P_{ji}$, $x_{P_i}$ and $x_{P_{ji}}$ are the centers of $P_i$ and $P_{ji}$, respectively, as shown in Fig. 9, and $\Delta x_{ji}$ denotes the Movement amount between $P_i$ and $P_{ji}$.

Next, determine the Movement amount of antecedent as follows.

$$
\Delta x_j = \frac{1}{s}\sum_{i=1}^{s}\varphi_i \Delta x_{ji},\ j=1,2,\cdots,n
\tag{24}
$$

where $\varphi_i$ is sign that reflects the proportional relationship between the $i^{th}$ input variable $x_i$ and the output variable z. If z is direct proportion to $x_i$, then $\varphi_i = 1$, else $\varphi_i = -1$. For some j, if $|\Delta x_j| > 1$, then we assume that the input information does not match the $j^{th}$ rule and exclude it from the calculation of the final reasoning result.

**Step 2:** Calculate the reasoning results for all fuzzy rules. Determine the Movement amounts of consequent from Movement amount $\Delta x_j, j=1,2,\cdots,n$ as follows.

$$\Delta z_j = F(\Delta x_j), \quad j=1,2,\cdots,n \tag{25}$$

where $F$ is a pre-defined function that reflects the Movement action on the consequent and $\Delta z_j$ is the Movement amount the consequent in $j^{th}$ rule. Move the consequent fuzzy set $Q_j$ to get the fuzzy reasoning result $Q'_j$ for $j^{th}$ rule.

$$Q'_j = \int_{Q'_j \in F(Z)} \mu_{Q'_j}(z) = \int_{Q_j \in F(Z)} \mu_{Q_j}(z)/(z+\Delta z_j), \quad j=1,2,\cdots,n \tag{26}$$

**Step 3:** Calculate the final fuzzy reasoning result $Q'$. The final fuzzy reasoning result $Q'$ is determined by the union of $Q'_j, j=1,2,\cdots,n$ shown as formula (27). The defuzzified crisp value of the fuzzy reasoning result can be calculated as formula (28).

$$Q' = \bigcup_{\substack{j=1 \\ |\Delta x_j| \leq 1}}^{n} Q'_j \tag{27}$$

$$z^0 = \int_{Q' \in F(Z)} z \cdot \mu_{Q'}(z)dz \Bigg/ \int_{Q' \in F(Z)} \mu_{Q'}(z)dz \tag{28}$$

For simplicity, the defuzzified value $z^0$ can be simply calculated as the arithmetic average of the centers of $Q'_j$.

$$z^0 = \tfrac{1}{n'} \sum_{\substack{i=1 \\ |\Delta x_j| \leq 1}}^{n'} (z_{Q_j} + \Delta z_j) \tag{29}$$

where, $n'$ denotes the number of rules participating in the computation of the reasoning result and $z_{Q_j}$ denotes the center of $Q_j$. For example, let us consider two fuzzy rules with three inputs and one output.

$R_1$: if $x_1$ is $P_{11}$, $x_2$ is $P_{12}$, $x_3$ is $P_{13}$ then $z$ is $Q_1$
$R_2$: if $x_1$ is $P_{21}$, $x_2$ is $P_{22}$, $x_3$ is $P_{23}$ then $z$ is $Q_2$
Input: $x_1$ is $P_1$, $x_2$ is $P_2$, $x_3$ is $P_3$
Conclusion: $z$ is $z^0 = ?$

The fuzzy reasoning process is as follows. Using formula (23), we compute the Movement amounts $\Delta x_{11}, \Delta x_{12}$ and $\Delta x_{13}$ between the antecedent fuzzy sets and the inputs for the rule $R_1$, where $x_{P_1}$, $x_{P_2}$, and $x_{P_3}$ are crisp input information, $x_{P_{11}}$, $x_{P_{12}}$ and $x_{P_{13}}$ are center of fuzzy sets $P_{11}, P_{12}$ and $P_{13}$, respectively.

$$\Delta x_{11} = \begin{cases} (x_{P_{11}} - x_{P_1})/r_{11}, & x_{P_1} \geq x_{P_{11}} \\ (x_{P_{11}} - x_{P_1})/l_{11}, & x_{P_1} < x_{P_{11}} \end{cases} \tag{30}$$

$$\Delta x_{12} = \begin{cases} (x_{P_{12}} - x_{P_2})/r_{12}, & x_{P_2} \geq x_{P_{12}} \\ (x_{P_{12}} - x_{P_2})/l_{12}, & x_{P_2} < x_{P_{12}} \end{cases} \tag{31}$$

$$\Delta x_{13} = \begin{cases} (x_{P_{13}} - x_{P_3})/r_{12}, & x_{P_3} \geq x_{P_{13}} \\ (x_{P_{13}} - x_{P_3})/l_{12}, & x_{P_3} < x_{P_{13}} \end{cases} \tag{32}$$

Likewise, compute the Movement amounts $\Delta x_{21}$, $\Delta x_{22}$, and $\Delta x_{23}$ for rule $R_2$. Next, from formula (15), compute the Movement amounts of antecedent $\Delta z_1$ and $\Delta z_2$.

$$\Delta z_1 = \tfrac{1}{3}(\pm \Delta x_{11} \pm \Delta x_{12} \pm \Delta x_{13}), \quad \Delta z_2 = \tfrac{1}{3}(\pm \Delta x_{21} \pm \Delta x_{22} \pm \Delta x_{23}) \tag{33}$$

Using formula (26) and formula (29), compute final defuzzified reasoning result $z^0$, where $z_{Q_1}$ and $z_{Q_2}$ are center of fuzzy sets $Q_1$ and $Q_2$, respectively.

$$z_1^0 = z_{Q_1} + \Delta z_1, \qquad z_2^0 = z_{Q_2} + \Delta z_2 \tag{34}$$

$$z^0 = \tfrac{1}{2}(z_1^0 + z_2^0) \tag{35}$$

Our method does not require complicated calculations and the computational complexity is $O(n)$. It does not include logical operations such as max or min, so mathematical analysis is convenient and easy to combine with other methods.

**3.2. T-S Fuzzy System based on MTP**

**3.2.1. T-S Fuzzy Reasoning Process based on MTP**

T-S fuzzy system model with $s$ inputs are given as formula (36).

$$R_j : \text{if } x_1 \text{ is } A_{j1} \; x_2 \text{ is } A_{j2} \cdots x_s \text{ is } A_{js} \text{ then } y_j = c_{j0} + c_{j1} \cdot x_1 + \cdots + c_{ji} \cdot x_i + \cdots + c_{js} \cdot x_s \tag{36}$$

where $A_{ji}$ is the i$^{\text{th}}$ antecedent triangular-shaped fuzzy set of the j$^{\text{th}}$ rule, $c_{i0}$ and $c_{ij}$ are the coefficients of the consequent linear function $(i = 1, 2, \cdots, s, \; j = 1, 2, \cdots, n)$. Without loss of generality, we suppose that inputs $x_i^0$ of fuzzy system are crisp values.

The fuzzy reasoning process based on MTP for the T-S fuzzy system is presented as follows (See Fig. 10)

.

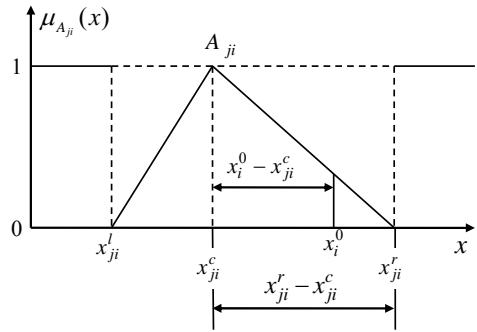

Fig. 10. Calculation of the Movement amounts with crisp input value

**Step 1**: Calculate the Movement amounts $d_{ji}, i=1,2,\cdots,s, j=1,2,\cdots,n$ between the center of the antecedent fuzzy set $A_{ji}$ and the input information $x_i^0$ by using formula (37). (See Fig. 10)

$$d_{ji} = \begin{cases} (x_i^0 - x_{ji}^c)/(x_{ji}^r - x_{ji}^c), & x_{ji}^r > x_i^0 \geq x_{ji}^c \\ (x_{ji}^c - x_i^0)/(x_{ji}^c - x_{ji}^l), & x_{ji}^l < x_i^0 < x_{ji}^c \\ 1, & x_i^0 \leq x_{ji}^l \text{ or } x_{ji}^r \leq x_i^0 \end{cases} \tag{37}$$

where $x_{ji}^c$, $x_{ji}^r$ and $x_{ji}^l$ denote the center, right endpoint, and left endpoint of the triangular-shaped fuzzy set $A_{ji}$ ($i=1,2,\cdots,s, j=1,2,\cdots,n$), respectively, and $x_i^0$ is the crisp input value corresponding to $i^{th}$ input variable as shown in Fig. 10.

**Step 2**: Calculate the Movement degree $d_j$ in the $j^{th}$ fuzzy rule for $j=1,2,\cdots,n$.

$$d_j = 1 - [d_{j1} \wedge d_{j2} \wedge \cdots \wedge d_{js}] \tag{38}$$

where, symbol $\wedge$ denotes the minimum or product operator.

**Step 3**: Construct a fuzzy rule subset $I_{act}$ to participate in the calculation of the fuzzy reasoning result.

$$I_{act} = \{j \mid d_j \geq \varepsilon, 1 \leq j \leq n\} \tag{39}$$

where $\varepsilon \in [0,1]$ is predefined threshold that allows rules that do not match the input to be excluded from the calculation of the fuzzy reasoning result.

**Step 4**: Calculate the final defuzzified crisp reasoning result $y^0$.

$$y^0 = \left(\sum_{j \in I_{act}} d_j \cdot y_j\right) \bigg/ \sum_{j \in I_{act}} d_j = \left(\sum_{j \in I_{act}} d_j \cdot (c_{j0} + c_{j1} \cdot x_1 + \cdots + c_{ji} \cdot x_j + \cdots + c_{js} \cdot x_s)\right) \bigg/ \sum_{j \in I_{act}} d_j \tag{40}$$

Our method is very similar to Sugeno's method [14] and Wang's method [20], and Hellendoorn's method [26], but our method has essentially distinct differences. It is shown in the next subsection 3.2.2, 3.2.3 and Section 4, respectively.

### 3.2.2. Comparison with Sugeno's and Our Method

Sugeno's method is a matching degree based fuzzy reasoning method applied to the T-S fuzzy model, in which the final reasoning result is calculated by weighted average based on the matching degree between the antecedent membership function and the input values. If the T-S fuzzy model has only one input, the matching degree of our method is equivalent to that of Sugeno's method. When the model has multiple inputs, matching degree of formula (30) is also consistent with the matching degree of Sugeno's method, since it also reflects the matching between the antecedent membership function and the input. Fig. 11 shows the relationship between Sugeno's method and proposed MTP method when fuzzy system has two inputs.

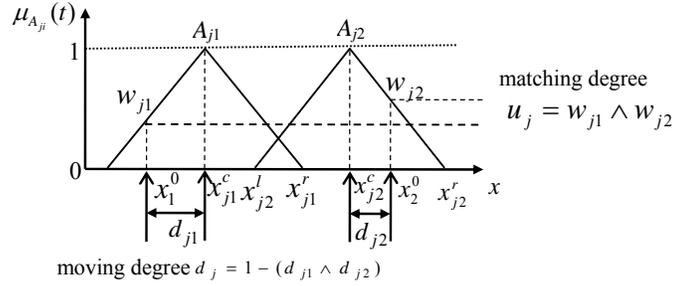

Fig. 11. Relationship between MTP method and Sugeno's method

As can see from Fig. 11, the Movement amount $d_{ji}$ of formula (37) is equivalent to membership $w_{ji}$, while Movement degree $d_j$ is equivalent to matching degree $u_j$ of CRI method. When the fuzzy set is triangular-shaped, the membership $w_{ji}$ of an crisp input information $x_i^0$ for antecedent triangular-shaped fuzzy set $A_{ji}$ is denoted as formula (41). From the formula (37), (38), and (41), it is clear that the larger matching degree of input information to fuzzy set, the smaller difference between input information and central point of fuzzy set, so the smaller Movement degree. In other words, both the matching degree and Movement degree reflect the degree of matching of the input value with the antecedent. Moreover, the final reasoning result is obtained by using weighted average based on the matching degree in both methods. However, our method is essentially different from the previous method because it examines the degree of matching between rules and input from the perspective of the Movement.

$$w_{ji} = \begin{cases} (x_{ji}^r - x_i^0)/(x_{ji}^r - x_{ji}^c), & x_{ji}^c \leq x_i^0 < x_{ji}^r \\ (x_i^0 - x_{ji}^l)/(x_{ji}^c - x_{ji}^l), & x_{lij} < x_i^0 < x_{cij} \\ 0, & x_i^0 \leq x_{ji}^l \text{ or } x_{ji}^r \leq x_i^0 \end{cases} \quad (41)$$

In addition, the Movement degree is not based membership function of input for the antecedent, and the form of the membership function has no significance in the reasoning process.

In the case of Sugeno's method, even if one input is not matched with the antecedent, the corresponding rule should be not participate in reasoning. Unlike, the proposed method allows such rules to participate in reasoning by appropriately setting the threshold value $\varepsilon$. By using the appropriate threshold, we can control the number of rules that participate in reasoning. In particular, since the proposed method does not require the calculation of the membership function for the input, the computation time required for the reasoning is much smaller than the previous methods.

### 3.2.3. Comparison with Distance-Type Fuzzy Reasoning and Our Method

Wang's method [20] is one kind of the distance-type fuzzy reasoning that derives the conclusion based on the distance between the consequent and input fuzzy set. The distance $d_j$ between the consequent and input fuzzy sets for the rule $R_j$, represented by formula (42), is calculated as follows. In formula (42) $A_{i0}, i = 1, 2, \cdots, n$ is the input fuzzy set corresponding to i$^{th}$ input variable. The crisp reasoning result $y^0$ is obtained by the following formula (43).

$$d_j = \sum_{i=1}^{n} d(A_{ji}, A_{i0}) \tag{42}$$

$$y^0 = \sum_{j=1}^{m} y_j \cdot \prod_{j \neq k} d_k \Big/ \sum_{j=1}^{m} \prod_{j \neq k} d_k \tag{43}$$

From formula (42) and formula (43), it can be known that Wang's method is much more complicated than our method in terms of computational complexity. Moreover, in the Wang's method, unnecessary distance calculation is performed even if all input fuzzy sets do not match with the corresponding antecedent (i.e., the corresponding rule does not affect the calculation of reasoning result). This way of the reasoning is incompatible with human thinking. If some information is obtained, the person does not use all his knowledge to obtain the result but uses only some appropriate knowledge. For MTP method, the Movement degree of the fuzzy rule is 0 and it is excluded from the calculation of the reasoning result in above case. Therefore, our MTP method is simpler than the distance-type method.

**5. Checking of [4]'s and Our Proposed Method**

In this paper, our proposed method is very similar to [4], which is different from [4] in several aspects. In this section, we check about [4]'s method and ours.

**4.1. Checking of "Constructing R with Powered Hedges" and Our Method**

In Section 4 of [4], author considered GMP using "powered hedge", where he used the following three conditions.

(1) "*if x is A then y is B*" is represented by a fuzzy relation R, which is built up from $A$ and $B$.

(2) When $\mu_{A'}(x) = \mu_A^m(x), m > 1$, then $A'$ is stronger than $A$; when $\mu_{A'}(x) = \mu_A^n(x)$, $0 < n < 1$, then $A'$ is weaker than $A$.

(3) $B'$ is calculated with the compositional rule of inference; $\mu_{B'}(y) = \max_x \min(\mu_{A'}(x), \mu_R(x, y))$.

In Section 2 of [4], powered hedges $A^+$ and $A^-$ are defined as formula (44) and (45).

$$A^+ = \int_{A \in F(X)} \mu_A^m(x)/x, \text{ for strengthening, i.e., } m > 1 \tag{44}$$

$$A^- = \int_{A \in F(X)} \mu_A^n(x)/x, \text{ for weakening, i.e., } 0 < n < 1 \tag{45}$$

In [4] the fuzzy relation $R$ is demonstrated by $\mu_R(x, y) = (1 - \mu_A(x)) \vee \mu_B(y)$, which does not satisfy the criteria, and also another fuzzy relation $R$: $\mu_R(x, y) = (\mu_A(x) \wedge \mu_B(y)) \vee (1 - \mu_A(x))$ does not satisfy. As shown Section 4 of [4], in case of using the powered hedge, fuzzy relation $R$ does not satisfy the criteria. In our method obtains $A' = \int_{A \in F(X)} \mu_{A^\alpha}(x)/x$ by the Transformation of $x_0$ by formula (11) in this paper and then obtain the fuzzy reasoning result as

$$B' = \int_{B \in F(Y)} \mu_{B^\beta}(y)/y = \int_{B \in F(Y)} \mu_{B^\alpha}(y)/y \quad \text{by Transformation index } \beta = \alpha \text{ without using fuzzy relation } R.$$

### 4.2. Checking of "Constructing R with Shifted Hedges" in [4] and Our Method

In Section 2 of [4], shifted hedges $A^+$ and $A^-$ are defined as follows.

$$A^+ = \int_{A \in F(X)} \mu_A(x-p)/x, \text{ for strengthening, i.e., } p > 0 \tag{46}$$

$$A^- = \int_{A \in F(X)} \mu_A(x-q)/x, \text{ for weakening, i.e., } q < 0 \tag{47}$$

where $A^+$ means strengthening, and $A^-$ weakening. In subsection 5.2 of [4], to show the inconsistencies of $R$, author used an example. The theorem 27 of [4] has shown and proved that the results given in formula (41) in [4] contradict the theory of the GMP with shifted hedges, hence, fuzzy relation $R$ is insufficient to represent "*if x is A then y is B*". Otherwise in our method, fuzzy relation is not used unlike Section 5 in [4], that is, as shown in formula (11) and (12), reasoning result $B'$ is obtained by Movement fuzzy set $B$ as Movement amount $\Delta y$ by $\Delta x$, Movement amount of $A$.

### 4.3. Checking of [4]'s " Functional Approach" and Our Method

First, consider a functional approach in [4]. In general, $B'$ is the function of $A$, $B$, and $A'$. The membership function of this function in [5] is as formula (48) and (49).

$$\mu_A(x) = \Gamma(x; a, b), \ \mu_B(y) = \Gamma(y; c, d), \ \mu_{A'}(x) = \Gamma(x; e, f), \text{ and } \mu_{B'}(y) = \Gamma(y; g, h) \tag{48}$$

$$A = \int_{\Gamma \in F(X)} \Gamma(x; a, b)/x, \quad B = \int_{\Gamma \in F(X)} \Gamma(y; c, d)/y \tag{49}$$

When $A$, $B$, and $A'$ are described using some $\Gamma$-function, $B'$ is also considered some $\Gamma$-function. It is difficult to give evidence for this statement, for details, see [5]. Their objective is to obtain the parameters g, h of reasoning result $B'$. For this, first, we decide the length of close interval $[g, h]$, and then obtain its center. Denote the fuzzy sets $A$ and $B$ of fuzzy rule as follows respectively.(Fig. 12) In Fig. 12, the length $l_A$ of $A$ is defined as $l_A = b - a$, and the center $m_A$ of $A$ as $m_A = \frac{1}{2}(a+b)$, and the measure of overlap $w(A,B)$ of two fuzzy sets is defined as formula (50). And the length of closed interval $[g, h]$ is calculated by the formula (51).

$$w(A, B) = [(a+b)-(c+d)]/[(b-a)+(d-c)] = \tfrac{1}{2}(m_A - m_B)/(l_A + l_B) \tag{50}$$

$$(b-a)/(d-c) = (f-e)/(h-g), \quad l_A/l_B = l_{A'}/l_{B'}, \quad l_{B'} = (l_{A'}/l_A) \cdot l_B \tag{51}$$

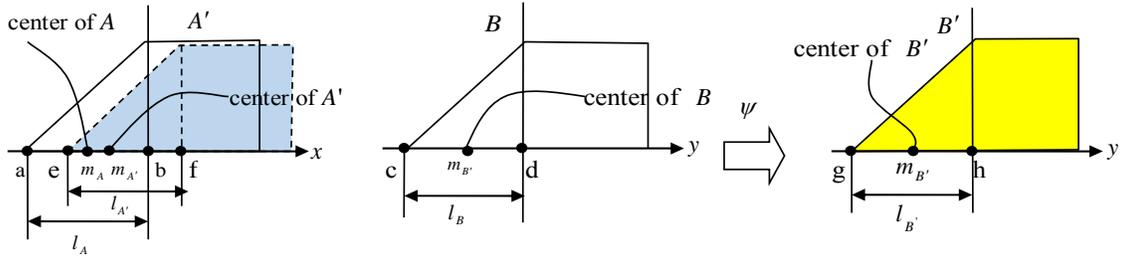

Fig. 12. A functional approach in [26]

In Fig. 12, the fuzziness of $A$ (resp. $A'$) implies $B$ (resp. $B'$). The center of closed interval $[g, h]$ is determined by the measure of overlap $w(A, B)$, denote $w$ for short. $w$ computes the difference between $A$ and $A'$, and works independent from the lengths of the shells of $A$ and $A'$. When $w = 0$, then the centers of $A$ and $A'$ coincide. And When $w > 0$, then $m_A > m_{A'}$. When $w \geq 1$ then $A'$ is "much weaker than" $A$. And when $w \leq -1$, then $A'$ is "much stronger than" $A$. We calculate $m_{B'}$ using a function $\Psi$ with $w$ and $l_{B'}$. If $w = 0$, then $m_A = m_{A'}$, thus, $m_B = m_{B'}$, so, $\Psi(w, l_{B'}) = l_B$. When $w = 1$, then $A'$ is "much weaker than" $A$, therefore $B' = $ unknown. By choosing that $h = \min(U)$, shell($B'$) falls outside the discourse of universe $U$, hence for all values $u \in U$, $\mu_{B'}(u) = 1$. This means the formula (52). When $w \leq -1$, then $m_A > m_{A'}$, therefore $B'$ is also "stronger than" $B$. Because $B'$ remains the same, strengthening $A'$ is useless. By assumption, let $m_{B'} = d$ in this case. Then the formula (53) is obtained. Using the function (54), the center of $B'$ is calculated. These observations deliver the function (54) to calculate the center of $B'$.

$$\Psi(1, l_{B'}) = \min(U) - \tfrac{1}{2} l_{B'} \tag{52}$$

$$\forall w \leq -1: \ \Psi(w, l_{B'}) = d \tag{53}$$

$$\Psi(w, l_B) = \tfrac{1}{2}\left[(\min(U) - \tfrac{1}{2} l_{B'} - c)w^2 + (\min(U) - \tfrac{1}{2} l_{B'} - d)w + (c + d)\right] \tag{54}$$

Conclusion of the fuzzy reasoning are as Fig. 12, formula (55) and (56).

$$g = \Psi(w(A, A'), l_{B'}) - \tfrac{1}{2} l_{B'}, \quad h = \Psi(w(A, A'), l_{B'}) + \tfrac{1}{2} l_{B'} \tag{55}$$

$$m_B = \tfrac{1}{2}(g + h) \tag{56}$$

Next, consider our proposed method. For this we explain through Fig. 13 according to Definition 2 in [4].

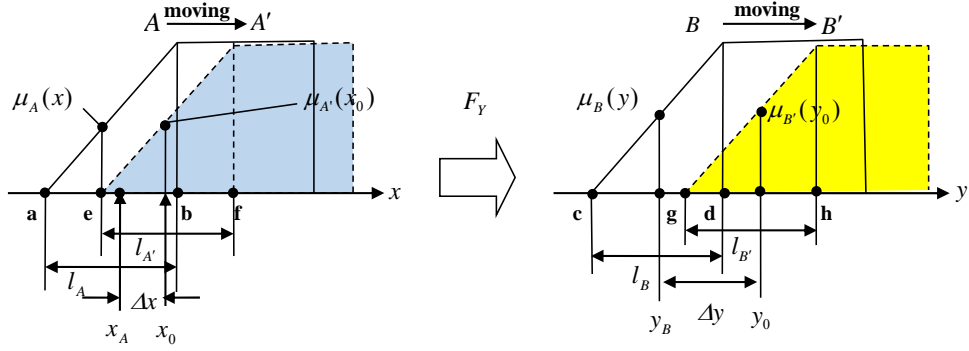

Fig. 13. Proposed method based on MTP

According to our method, the membership function of fuzzy set $A$ based on Definition 2 in [4] is calculated as follows.

$$\mu_A(x) = \int_{x \in [a,b]} (\frac{x-a}{b-a})/x = \int_{x \in [a,b]} (\frac{x-a}{l_A})/x \tag{57}$$

For the given crisp input information $x_0$, fuzzy set $A$ is moved to $A'$ as Movement amount $\Delta x = x_0 - x_A$ in Fig. 13 according to the idea of subsection 2.1 in this paper.

Concretely, if $\Delta x > 0$, then $A$ is moved to right hand, if $\Delta x < 0$, then $A$ is moved to left hand, and if $\Delta x = 0$, then $A$ is not moved. Therefore membership function of $A'$ from formula (5) and (11) based on MTP is calculated as follows.

$$\mu_{A'}(x) = F_X(\mu_A(x)) = \int_{A \in F(X)} \mu_A(x)/(x+\Delta x) = \int_{x \in [e,f]} (\frac{x_0-e}{f-e})/x = \int_{x \in [e,f]} (\frac{x_0-e}{l_{A'}})/x \tag{58}$$

Through Fig. 13 our method is explained for fuzzy reasoning shown in Fig. 12. By subsection 2.2, in formula (13), when $\Delta y = f(\Delta x) = \Delta x$, then fuzzy reasoning result $B'$ for FMP is calculated as follows.

$$\mu_{B'}(y) = F_Y(\mu_B(y)) = \int_{B' \in F(Y)} \mu_{B'}(y')/y' = \int_{B \in F(Y)} \mu_B(y)/(y+\Delta y) = \int_{B \in F(Y)} \mu_B(y)/(y+\Delta x) =$$
$$= \int_{y \in [c,d]} (\frac{y_B-c}{l_B})/(y+\Delta x) = \int_{y \in [g,h]} (\frac{y_0-g}{h-g})/(y+\Delta x) = \int_{y \in [g,h]} (\frac{y_0-g}{l_{B'}})/(y+\Delta x) \tag{59}$$

From the above consideration, we compare [4]'s functional approach and our proposed method.

The sameness of [4]'s and ours is as follows.

1) The membership function $\mu_A^m(x)$ and $\mu_A^n(x)$ in [4] presented as formula (44) and (45) are the same as $\mu_{A^\alpha}(x)$ in formula (11) in this paper, where $\alpha$ is Transformation index, $\alpha > 0$. When $\alpha > 1$ then $\mu_{A^\alpha}(x) = \mu_A^m(x)$, and when $0 < \alpha < 1$ then $\mu_{A^\alpha}(x) = \mu_A^n(x)$.

2) As shown in formula (11), (12), and (13), the Movement amount $\Delta x$ and $\Delta y$ in our method are equivalent to [4]'s shifted hedges $A^+$ and $A^-$ presented the formula (46) and (47), respectively.

3) For solving the fuzzy reasoning conclusion $B'$, information $[a, b]$, $[c, d]$, $[e, f]$ are all used in [4] as shown in formula (55) and (56), and our method as shown in formula (59).

The differences of [4]'s and ours are as follows.

1) [26] deals with a reasoning method of the case that input information is given as interval fuzzy set, our method deals with a reasoning method based on the observed crisp input. That is, in our method, new conclusion $B'$ is obtained by crisp input $x_0$, for example, as shown in formula (16) and (17), Movement amount $\Delta x$ is obtained from center $x_A$ of fuzzy set $A$ and input information $x_0$, reasoning result $B'$ is obtained by Movement amount $\Delta y$ based on $\Delta x$. In [4] new conclusion $B'$ is obtained based on $(x - p)$ and $(x - q)$, as shown in formula (46) and (47).

2) [4] deals with the case that two endpoints $g, h$ and center $m_{B'}$ of new fuzzy reasoning result $B'$ are obtained by using of the measure of overlap as shown in formula (53) ~ (56). Our method deals with the case that the new reasoning result $B'$ is obtained by the Movement amount $\Delta x$ and $\Delta y$, as shown in formula (13).

3) Our method deals with the fuzzification as shown in formula (16), (18), (20), and (58), and subsection 2.4, and defuzzification as shown in formula (5), (8), (28), (29), (35) and (40). And [4] deals with the measure of overlap for 2 fuzzy sets and its center as shown in formula (50), (55) and (56).

## 5. The Improved Learning Algorithm of Fuzzy Neural Network based on MTP

In order to verify the effectiveness of the proposed fuzzy reasoning method based on MTP, we carried out the learning experiments of the fuzzy neural network using MTP method and compared it with the Sugeno's method.

### 5.1. Configuration of pi-sigma Fuzzy Neural Network

The pi-sigma neural network is the neural network that has addition neuron and production neuron [4]. The output of pi-sigma neural network is as follows.

$$y^0 = \prod_{j=1}^{k} y_j = \prod_{j=1}^{k} \sum_{i=1}^{2} c_{ji} \cdot x_i \tag{60}$$

where $c_{ji}$ is the weight in neural network. As can be known from the structure of neural network, network output corresponds to the fuzzy reasoning output for the T-S system with two inputs, expressed as Fig. 14 and formula (60).

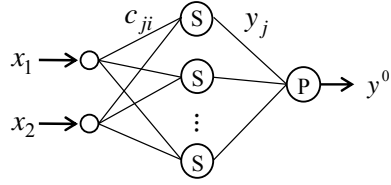

Fig. 14. pi-sigma neural network

Therefore, for T-S system, fuzzy reasoning method can be fully realized by using above neural structure. This neural network is called pi-sigma fuzzy neural network. In fuzzy neural networks, there are not only addition and multiplication operation but also fuzzy operation (e.g., maximum and minimum operation). The pi-sigma fuzzy neural network that realizes fuzzy reasoning method using MTP has showed in Fig. 15.

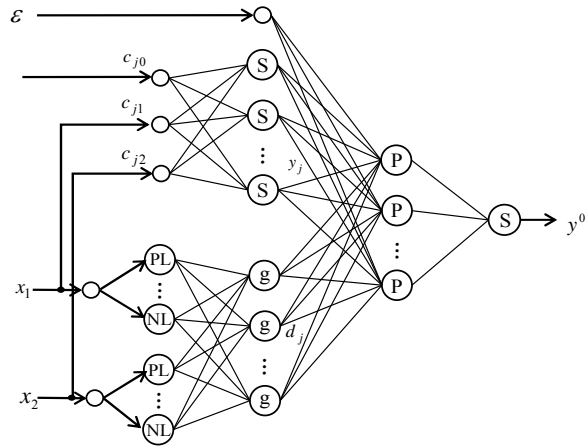

Fig. 15. pi-sigma fuzzy neural network by MTP

For convenience of neural network learning, all membership functions of every antecedent fuzzy sets are taken as Gaussian functions (61). The estimation function is defined as formula (62). Fig. 14 shows pi-sigma neural network with two input neurons and k hidden layers, where S denotes addition neuron and P denotes production neuron. In Fig. 15 neuron **g** calculates Movement degree from distance between central point of fuzzy set and input information. By using the threshold $\varepsilon$, the excitement level of neuron **P** is regulated. It means that only the necessary rules participate in the calculation when fuzzy reasoning is performed.

$$\mu_{A_{ji}}(x_i) = \exp\left[-(x_i - a_{ji})^2 / b_{ji}\right] \tag{61}$$

$$E = (y^d - y^0)^2 / 2 \tag{62}$$

In formula (61), $a_{ji}$ and $b_{ji}$ denote the center and width of fuzzy set $A_{ji}$, respectively, and in formula (62), $y^0$ is output of pi-sigma fuzzy neural network, and $y^d$ is the goal output of neural network. By the learning of fuzzy neural network, the centers of antecedent membership function and the coefficients of consequence linear function are decided, where the evaluation function E has the minimum value by using gradient method as formula (63), (64), and (65).

$$c_{ji}(t+1) = c_{ji}(t) - \eta \frac{\partial E}{\partial c_{ji}}\bigg|_{c_{ji}=c_{ji}(t)} \quad (63)$$

$$a_{ji}(t+1) = a_{ji}(t) - \eta \frac{\partial E}{\partial a_{ji}}\bigg|_{a_{ji}=a_{ji}(t)} \quad (64)$$

$$b_{ji}(t+1) = b_{ji}(t) - \eta \frac{\partial E}{\partial b_{ji}}\bigg|_{b_{ji}=b_{ji}(t)} \quad (65)$$

**5.2. Learning Experiment of Fuzzy Neural Network for Case Data Sets**

The precipitation data set used in this experiment is obtained from the precipitation data during 1952 to 1977 provided at China Tianjin city Weather Service as shown in Table 3.

**Table 3** Precipitation data set (mm)

| Year | Factor 1 | Factor 2 | Precipitation | Year | Factor 1 | Factor 2 | Precipitation |
|------|----------|----------|---------------|------|----------|----------|---------------|
| 1952 | 0.73 | -5.28 | 283 | 1965 | 0.46 | -14.68 | 348 |
| 1953 | -2.08 | 5.18 | 647 | 1966 | -2.31 | -1.36 | 644 |
| 1954 | -3.53 | 10.23 | 731 | 1967 | 0.20 | -5.43 | 431 |
| 1955 | -3.31 | 4.21 | 561 | 1968 | 3.46 | -19.85 | 179 |
| 1956 | 0.53 | -2.46 | 467 | 1969 | 0.08 | 8.59 | 615 |
| 1957 | 2.33 | 7.32 | 399 | 1970 | 1.46 | 7.26 | 433 |
| 1958 | -0.32 | -10.81 | 315 | 1971 | 0.24 | -1.10 | 401 |
| 1959 | -2.35 | 3.85 | 521 | 1972 | 0.89 | -16.94 | 206 |
| 1960 | -0.95 | 2.74 | 472 | 1973 | -0.50 | 10.46 | 639 |
| 1961 | -0.64 | 6.00 | 536 | 1974 | 2.15 | -10.06 | 418 |
| 1962 | 0.92 | 0.65 | 385 | 1975 | -0.89 | 12.11 | 570 |
| 1963 | 2.98 | -11.83 | 259 | 1976 | 1.40 | -6.26 | 415 |
| 1964 | -0.85 | -2.30 | 657 | 1977 | -0.59 | 7.15 | 796 |

The experiment were carried out on precipitation data and computer network security situation data, under the computational environment of R2012a MATLAB platform on a Window 8.1 (Intel(R) Core(TM) i7 CPU, 2.78 GHz processing, and 16 GB RAM). The 36 fuzzy rules have been made out for the experiment by setting 6 fuzzy sets {PL, PM, PS, NS, NM, NL} to every input variable $x_i\,(i=1,2)$.

$R_1$:    if $x_1$ is NL    and $x_2$ is NL    then $y_1 = c_{10} + c_{11}x_1 + c_{12}x_2$

$R_2$:    if $x_1$ is NL    and $x_2$ is NM    then $y_2 = c_{20} + c_{21}x_1 + c_{22}x_2$

                            …                        …

$R_{36}$:    if $x_1$ is PL    and $x_2$ is PL    then $y_{36} = c_{36\,0} + c_{36\,1}x_1 + c_{36\,2}x_2$

The comparison result of Sugeno's and the proposed method after 32 670 times of learning has shown in Table 4.

Table 4  Experiment results on precipitation data

| Year | Target value | Proposed method | | Sugeno's method | |
|---|---|---|---|---|---|
| | | Result | Error | Result | Error |
| 1952 | 283 | 416.38 | -133.38 | 400.55 | -117.55 |
| 1953 | 647 | 575.09 | 71.90 | 604.40 | 42.59 |
| 1954 | 731 | 781.51 | -50.51 | 708.96 | 22.03 |
| 1955 | 561 | 632.49 | -71.49 | 592.63 | -31.63 |
| 1956 | 467 | 416.23 | 50.76 | 358.45 | 108.54 |
| 1957 | 399 | 459.05 | -60.05 | 463.03 | -64.03 |
| 1958 | 315 | 315.73 | -0.730 | 376.05 | -61.05 |
| 1959 | 521 | 574.39 | -53.39 | 551.91 | -30.91 |
| 1960 | 472 | 483.82 | -11.82 | 457.71 | 14.28 |
| 1961 | 536 | 554.28 | -18.28 | 585.52 | -49.52 |
| 1962 | 385 | 363.65 | 21.34 | 243.62 | 141.37 |
| 1963 | 259 | 261.20 | -2.20 | 342.55 | -83.55 |
| 1964 | 657 | 618.27 | 38.72 | 461.29 | 195.70 |
| 1965 | 348 | 347.94 | 5.70 | 313.29 | 34.70 |
| 1966 | 644 | 644.36 | -0.36 | 480.93 | 163.06 |
| 1967 | 431 | 443.59 | -12.59 | 404.41 | 26.58 |
| 1968 | 179 | 178.92 | 7.30 | 146.19 | 32.80 |
| 1969 | 615 | 586.59 | 28.40 | 554.27 | 60.72 |
| 1970 | 433 | 494.92 | -61.92 | 473.37 | -40.37 |
| 1971 | 401 | 440.6 | -39.60 | 328.97 | 72.02 |
| 1972 | 206 | 206.00 | -8.72 | 237.53 | -31.53 |
| 1973 | 639 | 657.11 | -18.11 | 657.42 | -18.42 |
| 1974 | 418 | 393.30 | 24.69 | 381.56 | 36.43 |
| 1975 | 570 | 569.99 | 5.79 | 658.37 | -88.37 |
| 1976 | 415 | 392.19 | 22.80 | 404.70 | 10.29 |
| 1977 | 796 | 580.20 | 215.79 | 616.80 | 179.19 |

Table 5 shows the change of error according to the number of learning times. Prediction of network security situation evaluates security situation by learning data in the past and predicts the situation in the future. This method does prediction and studying by using fuzzy neural networks [10]. The output of pi-sigma neural network in experiment is calculated as formula (60).

**Table 5** Experiment results on precipitation data for different learning times

| Learning times | Sugeno's method | | Our proposed method | |
|---|---|---|---|---|
| | Learning time (s) | Learning error (%) | Learning time (s) | Learning error (%) |
| 100 | 0.7 | 94.710 | 0.6 | 68.114 |
| 500 | 4.9 | 77.685 | 1.4 | 11.870 |
| 1 000 | 9.8 | 67.205 | 2.1 | 10.803 |
| 1 500 | 14.7 | 59.911 | 3.5 | 9.370 |
| 2 000 | 19.6 | 53.567 | 4.2 | 9.253 |
| 4 000 | 38.5 | 39.594 | 8.4 | 9.531 |
| 8 000 | 70.7 | 25.427 | 16.8 | 8.582 |
| 10 000 | 97.3 | 23.184 | 20.3 | 8.516 |
| 15 000 | 147.0 | 20.478 | 31.5 | 8.832 |
| 18 000 | 175.7 | 19.163 | 37.1 | 8.078 |
| 20 000 | 192.5 | 18.340 | 41.3 | 8.640 |
| 25 000 | 237.3 | 16.899 | 51.8 | 7.887 |
| 30 000 | 284.2 | 15.799 | 61.6 | 8.318 |
| 32 670 | 516.6 | 15.260 | 67.2 | 7.911 |

Security situation data shows is in Table 6. And Table 7 shows the experiment results for different number of learning. For a simple comparison experiment, fuzzy level PL, PM and PS are chosen and 3 input variable are used. The fuzzy rules used in experiment on security situation by pi-sigma fuzzy neural network are as follows.

$R_1$: if $x_1$ is PL and $x_2$ is PL and $x_3$ is PL    then $y_1 = c_{10} + c_{11}x_1 + c_{12}x_2 + c_{13}x_3$
$R_2$: if $x_1$ is PL and $x_2$ is PL and $x_3$ is PM    then $y_2 = c_{20} + c_{21}x_1 + c_{22}x_2 + c_{23}x_3$
            …                                                                  …
$R_{27}$: if $x_1$ is PM and $x_2$ is PM and $x_3$ is PM    then $y_{27} = c_{27\,0} + c_{27\,1}x_1 + c_{27\,2}x_2 + c_{27\,3}x_3$

**Table 6** Measured data for security situation

| Factor 1 | Factor 2 | Factor 3 | Security situation value | Factor 1 | Factor 2 | Factor 3 | Security situation value |
|---|---|---|---|---|---|---|---|
| 1 | 2 | 7 | 13.792 | 2 | 1 | 8 | 16.354 |
| 1 | 6 | 9 | 14.783 | 9 | 4 | 2 | 94.707 |
| 5 | 1 | 9 | 37.333 | 8 | 4 | 8 | 77.354 |
| 5 | 2 | 9 | 37.748 | 6 | 2 | 3 | 48.992 |
| 7 | 6 | 8 | 62.803 | 8 | 3 | 7 | 77.110 |
| 4 | 2 | 1 | 29.414 | 6 | 4 | 5 | 49.447 |
| 8 | 6 | 6 | 77.858 | 9 | 4 | 6 | 94.408 |
| 6 | 9 | 3 | 50.577 | 4 | 1 | 6 | 28.408 |
| 4 | 2 | 6 | 28.822 | 4 | 8 | 4 | 30.328 |
| 5 | 2 | 1 | 38.414 | 2 | 6 | 7 | 17.827 |
| 1 | 1 | 4 | 13.500 | 2 | 4 | 9 | 17.333 |
| 3 | 4 | 7 | 22.378 | 1 | 3 | 7 | 14.110 |
| 1 | 5 | 2 | 14.943 | 5 | 6 | 7 | 38.827 |
| 1 | 8 | 9 | 15.162 | 2 | 7 | 8 | 17.999 |
| 9 | 1 | 1 | 94.000 | 7 | 8 | 6 | 63.237 |
| 9 | 3 | 7 | 94.110 | 4 | 9 | 6 | 30.408 |
| 3 | 9 | 9 | 23.333 | 1 | 4 | 6 | 14.408 |
| 4 | 7 | 5 | 30.093 | 3 | 9 | 1 | 24.000 |
| 3 | 4 | 9 | 22.333 | 3 | 7 | 6 | 23.054 |
| 5 | 7 | 1 | 39.646 | 1 | 7 | 3 | 15.223 |
| 4 | 7 | 7 | 30.024 | 1 | 1 | 1 | 14.000 |
| 7 | 8 | 9 | 63.162 | 5 | 7 | 7 | 39.024 |
| 2 | 4 | 4 | 17.500 | 8 | 1 | 7 | 76.378 |
| 1 | 8 | 4 | 15.328 | 4 | 6 | 8 | 29.803 |
| 6 | 7 | 1 | 50.646 | 2 | 4 | 7 | 17.378 |
| 4 | 1 | 1 | 29.000 | 7 | 4 | 8 | 62.354 |
| 9 | 4 | 1 | 95.000 | 1 | 4 | 1 | 15.000 |
| 6 | 2 | 1 | 49.414 | 6 | 6 | 5 | 49.897 |
| 3 | 1 | 8 | 21.354 | 4 | 1 | 9 | 28.333 |
| 5 | 4 | 5 | 38.447 | 3 | 2 | 4 | 21.914 |

The experiment result shows that the learning process of precipitation data was about 5.2 times faster and the learning process of security situation date was about 1.37 times faster compared to Sugeno's method. Computational time difference of the two methods will get bigger as fuzzy variables and rules are added. Learning accuracy of precipitation prediction has improved by 7.35% but prediction accuracy of security situation has decreased by 0.937%. The accuracy of learning will improve as the number of fuzzy partitions increases.

**Table 7** Experiment results on security situation for different learning times

| Learning times | Sugeno's method | | Proposed method | |
|---|---|---|---|---|
| | Learning time (s) | Learning error (%) | Learning time (s) | Learning error (%) |
| 100 | 2.8 | 25.558 | 2.10 | 37.169 |
| 500 | 11.9 | 12.001 | 9.10 | 18.366 |
| 1 000 | 23.8 | 10.231 | 18.2 | 9.148 |
| 1 500 | 34.3 | 8.919 | 26.6 | 6.695 |
| 2 000 | 44.8 | 7.854 | 35.7 | 5.772 |
| 4 000 | 90.3 | 5.375 | 69.3 | 2.968 |
| 8 000 | 179.9 | 3.757 | 135.8 | 4.299 |
| 10 000 | 225.4 | 3.523 | 168.7 | 4.090 |
| 15 000 | 338.1 | 3.072 | 251.3 | 3.565 |
| 18 000 | 448.0 | 2.850 | 307.3 | 2.876 |
| 20 000 | 450.8 | 2.727 | 335.3 | 2.914 |
| 25 000 | 569.1 | 2.476 | 417.2 | 3.065 |
| 30 000 | 674.1 | 2.279 | 492.8 | 2.642 |
| 32 670 | 751.1 | 2.191 | 533.4 | 2.383 |

The reason why the learning time is reduced for the proposed method is that it uses only the distance obtained from the membership function and the input information, and eliminates the unnecessary calculations in reasoning process by using threshold. In addition, Sugeno's method needs to calculate the membership functions with the form of exponential function while reasoning process. Unlike this, the proposed method based on MTP does not require such a time-consuming calculation. It can be easily known that the proposed method has higher efficiency compared to the previous method.

## 6. Conclusion

In this paper we proposed a new fuzzy reasoning method, called Movement and Transformation Principle of Fuzzy Reasoning (MTP), by deducing the Movement, Transformation, and Movement-Transformation relationship between the antecedent and input to the relationship between the consequent and conclusion, for FMP and FMT with single input and single output. The method works by first evaluating the Movement-Transformation relationship between the antecedent of fuzzy rule and input (the given observations), and then converting the consequent of fuzzy rule to the conclusion using the Movement and Transformation operations. A new model for the FMP and FMT representing the if-then fuzzy rule has proposed based on the compensating operation of the Movement-Transformation relations. The proposed method based on MTP is consistent with human thinking and satisfies logical reductive property. Our method was illustratively compared with Zadeh's, Sugeno's, Wang's, and Hellendoorn's method, respectively, which has some sameness, differences and independent property. The proposed method has applied to two fuzzy systems, i.e., Mamdani's one with s input 1 output, and T-S fuzzy neural network's one with s input linear function output, and then has compared with their previous methods. And the proposed method is computationally simple and does not involve strict logical operations, so it is easy to handle mathematically. Experimental evaluations of the proposed method through the fuzzy neural network indicate that it's learning accuracy and time performance are clearly improved and compared with previous Sugeno's method.

## References

1. S.M. Chen, A new approach to handling fuzzy decision-making problems, IEEE Transactions on Systems Man & Cybernetics 18 (1988) 1012–1016
2. S.M. Chen, A weighted fuzzy reasoning algorithm for medical diagnosis, Decision Support Systems 11 (1994) 37–43
3. G. Deng and Y. Jiang , Fuzzy reasoning method by optimizing the similarity of truth-tables, Information Sciences 288

(2014) 290–313

4. Hans Hellendoorn, The generalized modus ponens considered as a fuzzy relation, Fuzzy Sets and Systems 46(1): 2 (1992) 29–48

5. Hans Hellendoorn, Fuzzy logic and generalized modus ponens, Technical Report, University of Technology, Delft 88–92, 1988.

6. Y. Jin, J. Jiang and J. Zhu, Neural network based fuzzy identification and its application to modeling and control of complex systems, IEEE Transactions on Systems Man & Cybernetics 25 (1995) 990–997.

7. M. Luo and N. Yao, Triple I algorithms based on Schweizer–Sklar operators in fuzzy reasoning, International Journal of Approximate Reasoning 54 (2013) 640–652

8. M.X. Luo and Z. Cheng, Robustness of Fuzzy Reasoning Based on Schweizer-Sklar Interval-valued t-Norms, Fuzzy Information and Engineering 8 (2016) 183–198

9. H.W. Liu, Fully implicational methods for approximate reasoning based on interval-valued fuzzy sets, Journal of Systems Engineering and Electronics 21 (2010) 224–232

10. J. Lai，H. Wang, X. Liu and Y. Liang, A Quantitative Prediction Method of Network Security Situation Based on Wavelet Neural Network, International Symposium on Data 36 (2007) 197–202

11. E.H. Mamdani and S. Assilian, An experiment in linguistic synthesis with a fuzzy logic controller, International Journal of Human-Computer Studies 51 (1999) 1–13

12. M. Mizumoto and H.J. Zimmermann, Comparison of Fuzzy Reasoning Methods, Fuzzy Sets and Systems 8 (1982) 253–283

13. D. Pei, Triple I method for t-norm based logics, Fuzzy Systems and Mathematics 20 (2006) 1–7.

14. D. Pei, Unified full implication algorithms of fuzzy reasoning, Information Sciences 178 (2008) 520–30

15. M. Sugeno, Fuzzy identification of systems and its applications to modelling and control, IEEE Transactions on Systems Man and Cybernetics 15 (1985) 116–132

16. M. Sugeno, An introductory survey of fuzzy control, Information Sciences 36 (1985) 59–83

17. I.B. Turksen and Z. Zhao, An approximate analogical reasoning approach based on similarity measures, IEEE Transactions on Systems Man and Cybernetics 18 (1988) 1049–1056

18. I.B. Turksen and Z. Zhao, An approximate analogical reasoning scheme based on similarity measures and interval valued fuzzy sets, Fuzzy Sets and Systems 34 (1990) 323–346

19. D.G Wang, Y.P. Meng and H.X. Li, A fuzzy similarity inference method for fuzzy reasoning, Computers and Mathematics with Applications 56 (2008) 2445–2454

20. G.J. Wang, On the logic foundation of fuzzy reasoning, Information Sciences An International Journal 117 (1999) 47–88.

21. G.J. Wang, The full implication triple I method of fuzzy reasoning, Sci. China (Series E) 29 (1999) 43–53.

22. S. Wang, T. Tsuchiya and M. Mizumoto, Distance–Type Fuzzy Reasoning Method. Journal of Biomedical Fuzzy Systems Association 1 (1999) 61–78

23. D.S. Yeung and E.C.C. Tsang, Improved fuzzy knowledge representation and rule evaluation using fuzzy Petri nets and degree of subsethood, International Journal of Intelligent Systems 9 (1994) 1083–1100

24. D.S. Yeung and E.C.C. Tsang, A Comparative Study on Similarity–Based Fuzzy Reasoning Methods, IEEE Transactions on Systems Man & Cybernetics 27 (1997) 216–227

25. L.A. Zadeh, Fuzzy Sets, Information and Control 8 (1965) 338–353

26. L.A. Zadeh, Outline of new approach to the analysis of complex systems and decision processes. IEEE Transactions on Systems Man and Cybernetics 3 (1973) 28–33
27. Z.H. Zhao and Y.J. Li, Reverse triple I method of fuzzy reasoning for the implication operator $R_L$, Computers and Mathematics with Applications 53 (2007) 1020–1028

**Biography**

**CHUNG-JIN KWAK** received his B.S., and M.S. degrees, in Automation Engineering from **KIM IL SUNG** University, Pyongyang, from D P R of Korea in 2014, and 2017, respectively. His current research interests include Intelligence control using Fuzzy reasoning, Genetic Algorithm, and Neural Networks, and Ant Colony Algorithm.

**SON-IL KWAK** received his B.S., M.S., and Ph.D. degrees, all in Automatic control engineering, at **KIM IL SUNG** University, from D P R of Korea, Pyongyang, in 1990, 1994, and 2001, respectively. He was a visiting professor at the School of Electrical and Electronic Engineering, Beijing Telecommunication Technological University, P R of China, Beijing in 2006-2008. Currently he is a full professor at **KIM IL SUNG** University, Pyongyang. His current research interests include approximate reasoning using Fuzzy modelling, Fuzzy control, Fuzzy prediction, NN, GA, Membership cloud theory, Grey System prediction, Deep Learning, Rough set theory, ACO, PSO, and Computational Intelligence. He has co-authored a research of the theoretical and practical aspects in fuzzy reasoning and has the author or co-author of more than 10 published book chapters and papers in refereed international journals and conferences.

**DAE-SONG KANG** received his B.S., and M.S. degrees, in Information Processing from **KIM IL SUNG** University, Pyongyang, from D P R of Korea in 2006, and 2010, respectively. His current research interests include Intelligence System Design using Fuzzy Reasoning, Genetic Algorithm, and Neural Network.

**SONG-IL CHOE** received his B.S. degree, in Information Processing from **KIM IL SUNG** University, Pyongyang, from D P R of Korea in 2018. His current research interests include Intelligence Information Processing using Fuzzy Reasoning.

**JIN-UNG KIM** is an undergraduate at the School of Information, **KIM IL SUNG** University, Pyongyang, D P R of Korea. His major is Artificial Intelligence Technology. His currently research interests include Fuzzy set theory and Fuzzy reasoning technology, and Newral Network.

**HYOK-GI CHAE** received his B.S., M.S., Ph.D., and Prof. degrees, all in Information Processing engineering, at Kim Chaek Industry University, from D P R of Korea, Pyongyang, in 1980, 1983, 1989, and 1995, respectively. Currently he is a full academician at Pyongyang University of Computer Technology, Pyongyang. His current research interests include Expert System, Intelligence Decision Making System using Rough set theory, Ant Colony Optimization, Grey System, Particle Swarm Optimization, Genetic Algorithm, Neural Network, Fuzzy Set Theory, and Deep Learning. He has co-authored a research of the theoretical aspects in Intelligence Decision and has the author or co-author of published book chapters and papers in refereed international journals and conferences.